\documentclass{article}

\usepackage[preprint]{neurips_2020}

\usepackage[utf8]{inputenc} 
\usepackage[T1]{fontenc}    
\usepackage{hyperref}       
\usepackage{url}            
\usepackage{booktabs}       
\usepackage{amsfonts}       
\usepackage{nicefrac}       
\usepackage{microtype}      

\usepackage{amsmath}
\usepackage[pdftex]{graphicx}
\usepackage[flushleft]{threeparttable}
\usepackage{graphicx}

\usepackage{multirow}
\title{PAC-Bayesian Generalization Bounds for\\ MultiLayer Perceptrons}

\author{%
  Xinjie Lan, Xin Guo, Kenneth E. Barner\\ 
  Department of Electrical and Computer Engineering, University of Delaware \\
  \texttt{lxjbit@udel.edu} \\
}

\begin{document}

\maketitle

\begin{abstract}
We study PAC-Bayesian generalization bounds for Multilayer Perceptrons (MLPs) with the cross entropy loss.
Above all, we introduce probabilistic explanations for MLPs in two aspects: (i) MLPs formulate a family of Gibbs distributions, and (ii) minimizing the cross-entropy loss for MLPs is equivalent to Bayesian variational inference, which establish a solid probabilistic foundation for studying PAC-Bayesian bounds on MLPs.
Furthermore, based on the Evidence Lower Bound (ELBO), we prove that MLPs with the cross entropy loss inherently guarantee PAC-Bayesian generalization bounds, and minimizing PAC-Bayesian generalization bounds for MLPs is equivalent to maximizing the ELBO.
Finally, we validate the proposed PAC-Bayesian generalization bound on benchmark datasets.
\end{abstract}

\section{Introduction}

To explain why Deep Neural Networks (DNNs) with Stochastic Gradient Descent (SGD) optimization can achieve impressive generalization performance \cite{kawaguchi2017generalization}, a generalization bound should satisfy the five criteria: (i) small and non-vacuous \cite{dziugaite2017computing, zhou2018non}; (ii) taking into account SGD optimization because DNNs not incorporate any explicit regularization \cite{allen2019learning, hoffer2017train, li2018learning}; (iii) showing the same architecture dependence as the generalization error, e.g., it should become smaller as the architecture of DNNs becomes complex \cite{generalization_regularization1, poggio2017and, regularization5}; (iv) showing the same sample size dependence as the generalization error, i.e., it should become smaller as the training sample size becomes larger \cite{nagarajan2019uniform, generalization_regularization1}; and (v) taking into account the statistical variations of samples and labels, especially random labels \cite{generalization_regularization}.

PAC-Bayesian theory provides a canonical probabilistic framework for studying generalization bounds \cite{mcallester1999some, mcallester2003pac, catoni2007pac, germain2009pac, alquier2016properties, nagarajan2019deterministic}.
However, due to the extremely complicated architecture of DNNs, most existing works avoid directly formulating the prior distribution of the hypothesis expressed by DNNs and the posterior distribution given training samples through either relaxing the KL-divergence involved PAC-Bayesian theory \cite{neyshabur2017pac, zhou2018non, lever2013tighter} or approximating the distributions under some assumptions \cite{letarte2019dichotomize, dziugaite2017computing, london2017pac, ambroladze2007tighter}, which make existing PAC-Bayesian generalization bounds difficult to satisfy the above criteria\footnote{We provide more discussions about PAC-Bayesian generalization bounds on DNNs in Appendix \ref{related_works}.} \cite{nagarajan2019uniform, generalization_regularization1}.
Therefore, they cannot fully explain the generalization performance of DNNs.

In this paper, we study PAC-Bayesian generalization bounds on fully connected feedforward neural networks, namely MultiLayer Perceptrons (MLPs).
Above all, we attempt to introduce explicitly probabilistic explanations for MLPs in three aspects: (i) we define a probability space for a fully connected layer based on Gibbs distribution \cite{energy_learning, Geman, Boltzmann_machine}; (ii) we prove the entire architecture of MLPs corresponding to a family of Gibbs distributions; and (iii) we prove the equivalence between the cross entropy loss minimization and Bayesian variational inference \cite{VI, SVI, wainwright2008graphical}.
In contrast to previous works using relaxation or approximation, the explicitly probabilistic explanations for MLPs establish a solid probabilistic foundation for studying PAC-Bayesian generalization bounds on MLPs.

Inspired by the previous work \cite{germain2016pac}, we study PAC-Bayesian generalization bounds on MLPs from the perspective of Bayesian theory.
Since minimizing the cross entropy loss of MLPs is equivalent to Bayesian variational inference, we prove that MLPs with the cross entropy loss inherently guarantee PAC-Bayesian generalization bounds based on the Evidence Lower Bound (ELBO) derived from Bayesian variational inference \cite{VI, SVI}.
To the best of our knowledge, this is the first time theoretically guaranteeing the PAC-Bayesian generalization performance for MLPs.
Finally, we derive a novel PAC-Bayesian generalization bound and validate the bound on benchmark datasets.

\section{PAC-Bayesian Theory}

We assume that $\mathcal{D}$ is a data generating distribution and $\mathcal{S} = \{(x_i, y_i)\}_{i=1}^{n} \in (\mathcal{X} \times \mathcal{Y})^{n}$ is composed of $n$ \textit{i.i.d.}\ samples generated from $\mathcal{D}$, i.e., $\mathcal{S} \sim \mathcal{D}^n$.
We define $\mathcal{H}$ as a set of hypothesis $h: \mathcal{X} \rightarrow \mathcal{Y}$ and the loss function as $\ell: \mathcal{H}\times \mathcal{X} \times \mathcal{Y} \rightarrow \mathbb{R}$, thus the empirical risk on the samples $\mathcal{S}$ and the generalization error on the distribution $\mathcal{D}$ are
\begin{equation} 
\hat{\mathcal{L}}^{\ell}_{\mathcal{S}}(h) = \frac{1}{n}\sum_{i=1}^n\ell(h, x_i,y_i); \mathcal{L}^{\ell}_{\mathcal{D}}(h) = \underset{(x,y)\sim\mathcal{D}}{\boldsymbol{E}}\ell(h,x,y).
\end{equation}

The PAC-Bayesian theory establishes the connection between the \textit{probably approximately correct}  generalization bound ${\boldsymbol{E}_{h \sim \rho}}\mathcal{L}^{\ell}_{\mathcal{D}}(h) = {\boldsymbol{E}_{h \sim \rho}}{\boldsymbol{E}_{(x,y)\sim\mathcal{D}}} \ell(h,x,y)$ and the expectation of the empirical risk ${\boldsymbol{E}_{h \sim \rho}}\hat{\mathcal{L}}^{\ell}_{\mathcal{S}}(h)$, where $\rho$ denotes the posterior distribution of $h$ given $\mathcal{S}$.

\paragraph{Theorem 1:} PAC-Bayesian bounds on the generalization risk (Germain \textit{et al.}\cite{germain2016pac}).

Given a distribution $\mathcal{D}$ over $\mathcal{X} \times \mathcal{Y}$, a hypothesis set $\mathcal{H}$, a loss function $\ell': \mathcal{F}\times \mathcal{X} \times \mathcal{Y} \rightarrow [0,1]$, a prior distribution $\pi$ over $\mathcal{H}$, a real number $\delta \in (0,1]$, and a real number $\beta > 0$, with probability at least $1-\delta$ over the samples $\mathcal{S} \sim \mathcal{D}^n$, we have
\begin{equation} 
\forall \rho \text{ on } \mathcal{H}: \underset{{h \sim \rho}}{\boldsymbol{E}}\mathcal{L}^{\ell'}_{\mathcal{D}}(h) \leq \frac{1}{1-e^{-\beta}}[1-e^{-\beta {\boldsymbol{E}_{h \sim \rho}}\hat{\mathcal{L}}^{\ell'}_{\mathcal{S}}(h)-\frac{1}{n}(\text{KL}[\rho || \pi]+ln\frac{1}{\delta})}].
\end{equation}
We can extend it to arbitraray bounded loss, i.e., $\ell: \mathcal{F}\times \mathcal{X} \times \mathcal{Y} \rightarrow [a, b]$, where $[a,b] \in \mathbb{R}$ through defining $\beta:= b -a $ and $\ell'(h, x, y) := (\ell(h, x, y)-a)/(b-a) \in [0,1]$.
\begin{equation} 
\label{pac_bayesian}
\forall \rho \text{ on } \mathcal{H}: \underset{{h \sim \rho}}{\boldsymbol{E}}\mathcal{L}^{\ell}_{\mathcal{D}}(h) \leq \frac{b-a}{1-e^{a-b}}\{1-e^{[-{\boldsymbol{E}_{h \sim \rho}}\hat{\mathcal{L}}^{\ell}_{\mathcal{S}}(h)-\frac{1}{n}(\text{KL}[\rho || \pi]+ln\frac{1}{\delta})+a]}\}.
\end{equation}
Formula \ref{pac_bayesian} indicates that minimizing the generalization bound is equivalent to minimizing
\begin{equation}
\label{pac_equivalence}
n{\boldsymbol{E}_{h \sim \rho}}\hat{\mathcal{L}}^{\ell}_{\mathcal{S}}(h)+\text{KL}[\rho || \pi].
\end{equation}

\begin{figure}
\centering
\includegraphics[scale=0.65]{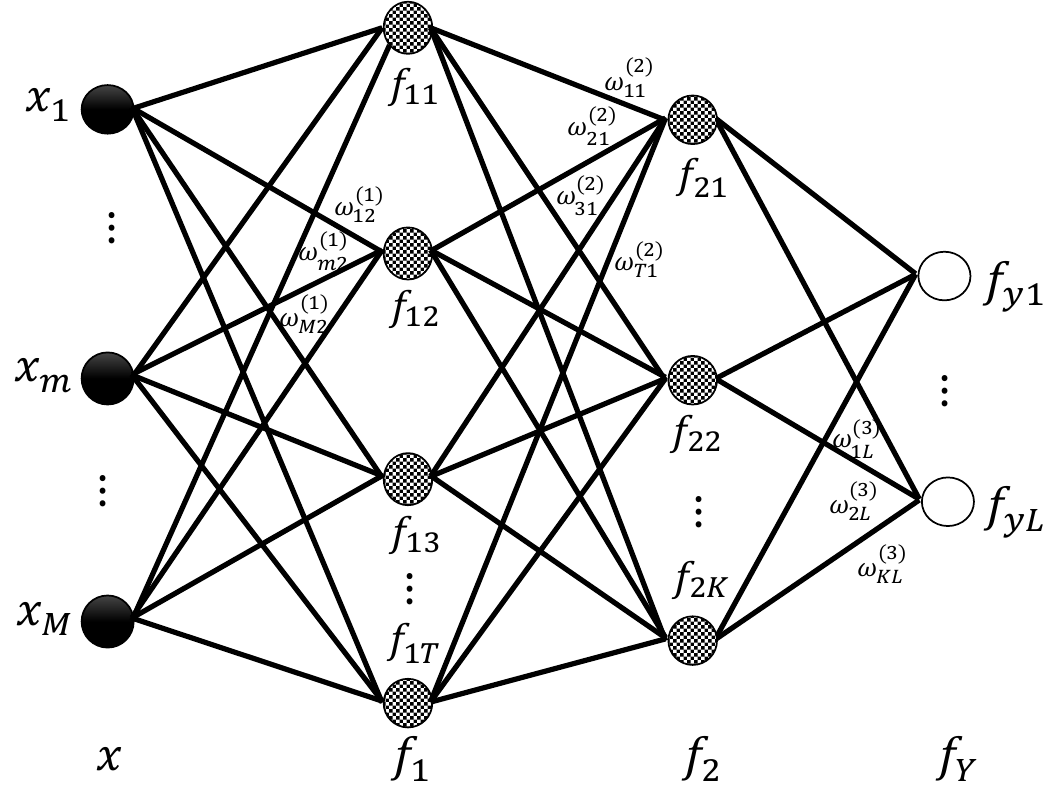}
\caption{
The $\text{MLP} = \{{x; f_1; f_2; f_Y}\}$. The input layer ${x}$ has $M$ nodes, and ${f_1}$ has $T$ neurons, i.e., ${f_1} = \{f_{1t} = \sigma_1[g_{1t}({x})]\}_{t=1}^{T}$, where $g_{1t}(x) = \sum_{m=1}^M\omega^{(1)}_{mt} \cdot x_{m} + b_{1t}$ is the $t$th linear function with $\omega^{(1)}_{mt}$ being the weight of the edge between $x_{m}$ and $f_{1t}$ and $b_{1t}$ being the bias, and $\sigma_1(\cdot)$ is a non-linear activation function, e.g., ReLU function.
Similarly, ${f_{2}} = \{f_{2k}=\sigma_2[g_{2k}({f_1})]\}_{k=1}^K$ has $K$ neurons, where $g_{2k}({f_1}) = \sum_{t=1}^T\omega^{(2)}_{tk} \cdot f_{1t} + b_{2k}$.
The output layer ${f_Y}$ is the softmax, thus $f_{yl} = \frac{1}{{Z(f_2)}}\text{exp}[g_{yl}(f_2)]$, where $g_{yl}(f_2)=\sum_{k=1}^K\omega^{(3)}_{kl} \cdot f_{2k} + b_{yl}$ and $Z(f_2) = \sum_{l=1}^L\text{exp}[g_{yl}({f_2})]$.
}
\label{fig_mlp}
\end{figure}

To facilitate subsequent discussions, we define $X$, $Y$, and $F_Y$ as the random variables of the sample $x_i \in \mathcal{X}$, the label $y_i \in \mathcal{Y}$, and the hypothesis $h(x_i) \in \mathcal{Y}$, respectively. 
Therefore, $\pi := P(F_Y|X)$ and $\rho := P(F_Y|X,Y)$ denote the prior distribution and the posterior distribution of $h$, respectively. Based on Bayes' rule, we can formulate the connection between $P(F_Y|X)$ and $P(F|X,Y)$ as
\begin{equation} 
P(F_Y|X,Y)  = \frac{P(F_Y|X)P(Y|F_Y,X)}{P(Y|X)},
\end{equation}
where $P(Y=y_i|F_Y,X=x_i)$ is the likelihood of $y_i \in \mathcal{Y}$ given $h$ and $x_i \in \mathcal{X}$, and the evidence can be formulated as $P(Y|X) = \sum_{l \in \mathcal{Y}}P(F_Y=l|X)P(Y|F_Y=l,X)$\footnote{We only consider the discrete case, because we prove $F_Y$ as a discrete random variable in the next section.}.

In the context of supervised classification based on MLPs, $\mathcal{Y} = \{1, \cdots, L\}$ consists of finite $L$ labels, and $P_{F_Y|X}(l|x_i)$\footnote{For simplicity, $P_{F_Y|X}(l|x_i) := P(F_Y=l|X=x_i)$.} denotes the prior probability of $x_i$ belonging to class $l$ based on the $h$ expressed by MLPs.
We define $P^*(F_Y|X)$ as the truly prior distribution of the true $h^*$.
Given $P^*(F_Y|X)$ and $P(Y|F_Y,X)$, $P^*(F_Y|X,Y)$ is the truly posterior distribution with the one-hot format, i.e., $\forall l \in \mathcal{Y}$, $P^*_{F_Y|X,Y}(l|x_i, y_i)  = 1$ if $l = y_i$; otherwise $P^*_{F_Y|X,Y}(l|x_i,y_i) = 0$. 
We aim to learn an optimal MLP from $\mathcal{S}$ such that it approaches $P^*(F_Y|X,Y)$ as close as possible.
In this paper, we use the $\text{MLP} = \{{x; f_1; f_2; f_Y}\}$ (Figure \ref{fig_mlp}) for most theoretical derivations unless otherwise specified.

\section{Probabilistic Explanations for Multilayer Perceptrons}

In this section, we define the probability space $(\Omega, \mathcal{T}, Q_F)$ for a fully connected layer, thereby deriving that the entire architecture of the MLP formulates a family of Gibbs distributions $Q(F_Y|X)$. 
Moreover, we prove that if $\ell$ is the cross entropy, minimizing $\hat{\mathcal{L}}^{\ell}_{\mathcal{S}}(h)$ can be explained as Bayesian variational inference, and it is equivalent to maximizing $Q_{\mathcal{S}}(F_Y|X) = \prod_{i=1}^n Q_{F_Y|X}(y_i|x_i)$.  

\paragraph{Definition 1:} the probability space $(\Omega, \mathcal{T}, Q_F)$ for a fully connected layer.

Given a fully connected layer ${f}$ with $T$ neurons $\{f_t = \sigma[g_{t}({x_i})]=\sigma(\sum_{m=1}^M\omega_{mt}\cdot x_{im} + b_t)\}_{t=1}^T$, where $\sigma(\cdot)$ is an activation function, e.g, ReLU function, $x_i = \{x_{im}\}_{m=1}^M \in \mathcal{X}$ is the input of ${f}$, and $g_t(\cdot)$ is the $t$th linear filter with $\omega_{mt}$ being the weight and $b_t$ being the bias, let inputting ${x_i}$ into ${f}$ be an experiment, we define the probability space $(\Omega, \mathcal{T}, Q_F)$ as follows.

First, the sample space $\Omega$ consists of $T$ possible outcomes $\{\boldsymbol{\omega}_t\}_{t=1}^T$, where $\boldsymbol{\omega}_t = \{\omega_{mt}\}_{m=1}^M$.
In terms of machine learning, a possible outcome $\boldsymbol{\omega}_t$ defines a potential feature of ${x_i}$.
Since scalar values cannot describe the dependence within ${x_i}$, we do not take into account $b_t$ for defining $\Omega$.
Second, we define the event space $\mathcal{T}$ as the $\sigma$-algebra.
For example, if ${f}$ has $T = 2$ linear filters and $\Omega = \{\boldsymbol{\omega}_1, \boldsymbol{\omega}_2\}$, thus ${\textstyle \mathcal{T} = \{\emptyset, \{\boldsymbol{\omega}_1\}, \{\boldsymbol{\omega}_2\}, \{\{\boldsymbol{\omega}_1, \boldsymbol{\omega}_2\}\}}$ indicates that none of outcomes, one of outcomes, or both outcomes could happen in the experiment.
Third, we prove the probability measure $Q_F$ as a Gibbs distribution to quantify the probability of possible outcomes $\{\boldsymbol{\omega}_t\}_{t=1}^T$ occurring in ${x_i}$.
\begin{equation} 
\label{Gibbs_f}
Q_F(\boldsymbol{\omega}_t) = \frac{1}{Z_F(x_i)}\text{exp}(f_{t}) = \frac{1}{Z_F(x_i)}\text{exp}[\sigma(\langle \boldsymbol{\omega}_t, {x_i} \rangle + b_t)],
\end{equation}
where $\langle \cdot, \cdot \rangle$ is the inner product, $Z_F(x_i) = \sum_{t=1}^T\text{exp}(f_{t})$ is the partition function only depending on $x_i$, and the energy function $E_{\boldsymbol{\omega}_t}({x_i})$ equals the negative activation, i.e., $E_{\boldsymbol{\omega}_t}({x_i}) = -f_t$.
Notably, the higher activation $f_t$ indicates the lower energy $E_{\boldsymbol{\omega}_t}({x_i})$, thus $\boldsymbol{\omega}_t$ has the higher probability of occurring in ${x_i}$.
The detailed proofs and supportive simulations are presented in Appendix \ref{prob_space}\footnote{We recommend reading the supplementary file connecting the paper and the appendix together}.

Based on $(\Omega, \mathcal{T}, Q_F)$ for the fully connected layer ${f}$, we can explicitly define the corresponding random variable $F: \Omega \rightarrow E$.
Specifically, since $\Omega = \{\boldsymbol{\omega}_t\}_{t=1}^T$ includes finite $T$ possible outcomes, $E$ should be a discrete measurable space and $F$ is a discrete random variable. 
For example, $Q(F = t)$ indicates the probability of the $t$th outcome $\boldsymbol{\omega}_t$ occurring in ${x_i}$.
Furthermore, the probability space of a fully connected layer enables us to derive probabilistic explanations for MLPs.

\pagebreak
\paragraph{Theorem 2:} the $\text{MLP} = \{{x; f_1;f_2; f_Y}\}$ corresponds to a family of Gibbs distributions 
\begin{equation} 
\label{pdf_posterior1}
Q_{F_Y|X}(l|x_i) = \frac{1}{Z_{\text{MLP}}(x_i)}\text{exp}[-E_{yl}(x_i)] = \frac{1}{Z_{\text{MLP}}(x_i)}\text{exp}[{f_{yl}(f_2(f_1({x_i})))}],
\end{equation}
where $E_{yl}(x_i) = -{f_{yl}(f_2(f_1(x_i)))}$ is the energy function of $l \in \mathcal{Y}$ given $x_i$, and the partition function $Z_{\text{MLP}}(x_i) = \sum_{l=1}^L \sum_{k=1}^K \sum_{t=1}^T Q(F_Y, F_2, F_1|X=x_i)$.

\textit{Proof:} Definition 1 implies $\omega_{mt}$ defining $\Omega$, thus $\Omega$ would be fixed if not considering training.
Therefore, ${F_{i+1}}$ is entirely determined by ${F_{i}}$ in the MLP, where $F_i$ denotes the random variable of $f_i$, and the MLP forms a Markov chain ${X} \boldsymbol{\leftrightarrow} {F_1} \boldsymbol{\leftrightarrow} {F_2} {\leftrightarrow} {F_Y}$.
As a result, given a sample $x = x_i$, the conditional distribution $Q(F_Y|X)$ corresponding to the MLP can be derived as 
\begin{equation}
Q_{F_Y|X}(l|x_i) = \sum_{k=1}^K \sum_{t=1}^T Q_{F_Y, F_2, F_1|X}(l, k, t|x_i) = \frac{1}{Z_{\text{MLP}}(x_i)}\text{exp}[{f_{yl}(f_2(f_1({x_i})))}].
\end{equation}
The derivation is presented in Appendix \ref{posterior_MLP}.
The energy function $E_{yl}(x_i) = -{f_{yl}(f_2(f_1(x_i)))}$ implies that $Q_{F_Y|X}(l|x_i)$ is entirely determined by the architecture of the MLP.
In other words, the MLP formulates a family of Gibbs distributions $Q(F_Y|{X})$.

\paragraph{Corollary 1:} Given the samples $\mathcal{S}$ and the $\text{MLP} = \{{x; f_1;f_2; f_Y}\}$, if $\ell$ is the cross entropy, then minimizing $\hat{\mathcal{L}}^{\ell}_{\mathcal{S}}(h)$ is equivalent to maximizing $Q_{\mathcal{S}}(F_Y|X)$.
\begin{equation} 
\underset{h \in \mathcal{H}}{\text{min}}\hat{\mathcal{L}}^{\ell}_{\mathcal{S}}(h) \Rightarrow \underset{Q(F_Y|X)}{\text{max}}\frac{1}{n}\text{log}Q_{\mathcal{S}}(F_Y|X),
\end{equation}
where $Q_{\mathcal{S}}(F_Y|X) = \prod_{i=1}^n Q_{F_Y|X}(y_i|x_i)$.

\textit{Proof:} since the cross entropy $H[P^*(F_Y|X, Y),Q(F_Y|X)] = \underset{P^*(F_Y|X, Y)}{-\boldsymbol{E}}[\text{log}Q(F_Y|X)]$, we have
\begin{equation}
\label{cross_entropy}
\hat{\mathcal{L}}^{\ell}_{\mathcal{S}}(h) = -\frac{1}{n}\sum_{i=1}^n H[P^*(F_Y|X, Y),Q(F_Y|X)] = -\frac{1}{n}\sum_{i=1}^n\sum_{l=1}^LP^*_{F_Y|X, Y}(l|x_i,y_i)\text{log}Q_{F_Y|X}(l|x_i).
\end{equation}

Since $P^*(F_Y|X, Y)$ is one-hot, namely that if $l = y_i$, then $P^*_{F_Y|X, Y}(l|x_i, y_i)  = 1$, otherwise $P^*_{F_Y|X, Y}(l|x_i, y_i) = 0$,  we can simplify Equation (\ref{cross_entropy}) as
\begin{equation}
\hat{\mathcal{L}}^{\ell}_{\mathcal{S}}(h) = -\frac{1}{n}\sum_{i=1}^n\text{log}Q_{F_Y|X}(y_i|x_i) = -\frac{1}{n}\text{log}Q_{\mathcal{S}}(F_Y|X).
\end{equation}
Therefore, minimizing $\hat{\mathcal{L}}^{\ell}_{\mathcal{S}}(h)$ is equivalent to maximizing the likelihood $Q_{\mathcal{S}}(F_Y|X)$.
In addition, based on Theorem 2 (Equation \ref{pdf_posterior1}), we can extend $\hat{\mathcal{L}}^{\ell}_{\mathcal{S}}(h)$ as 
\begin{equation}
\label{energy_minimization}
n\hat{\mathcal{L}}^{\ell}_{\mathcal{S}}(h) = -\sum_{i=1}^n\text{log}Q_{F_Y|X}(y_i|x_i) = \sum_{i=1}^n[E_{yy_i}(x_i) + \text{log}Z_{\text{MLP}}(x_i)].
\end{equation}
Recall that $Z_{\text{MLP}}(x_i) = \sum_{l=1}^L \sum_{k=1}^K \sum_{t=1}^T Q(F_Y, F_2, F_1|X=x_i)$ sums up all the cases of $F_Y$, thus $\text{log}Z_{\text{MLP}}(x_i)$ can be viewed as a constant with respect to $Q(F_Y|X)$.

\paragraph{Theorem 3:} 
given the samples $\mathcal{S}$ and the $\text{MLP} = \{{x; f_1;f_2; f_Y}\}$, if $\ell$ is the cross entropy, then minimizing $\hat{\mathcal{L}}^{\ell}_{\mathcal{S}}(h)$ can be explained as Bayesian variational inference.
\begin{equation} 
\underset{h \in \mathcal{H}}{\text{min}}\hat{\mathcal{L}}^{\ell}_{\mathcal{S}}(h) \Rightarrow \underset{Q(F_Y|X)}{\text{min}}\text{KL}[Q_{\mathcal{S}_x}(F_Y|X)||P^*_{\mathcal{S}}(F_Y|X, Y)],
\end{equation}
where $Q_{\mathcal{S}_x}(F_Y|X) = \prod_{i=1}^n Q(F_Y|X=x_i)$ and $P^*_{\mathcal{S}}(F_Y|X, Y) = \prod_{i=1}^nP^*(F_Y|X=x_i, Y=y_i)$.

\textit{Proof:} based on the property $H(P,Q) = H(P) + \text{KL}[P||Q]$, we can derive $\hat{\mathcal{L}}^{\ell}_{\mathcal{S}}(h)$ as
\begin{equation} 
\hat{\mathcal{L}}^{\ell}_{\mathcal{S}}(h) = \frac{1}{n}\sum_{i=1}^n H[{ P^*(F_Y|X =x_i, Y=y_i)}] + \text{KL}[{ P^*(F_Y|X =x_i, Y=y_i)||Q(F_Y|X=x_i)}].
\end{equation}
Since $P^*_{F_Y|X, Y}(l|x_i, y_i)$ is one-hot, namely that if $l = y_i$, $P^*_{F_Y|X, Y}(l|x_i, y_i)  = 1$, otherwise $P^*_{F_Y|X, Y}(l|x_i, y_i) = 0$, $H[P^*(F_Y|X=x_i, Y=y_i)] =0$, thus we can simplify $\hat{\mathcal{L}}^{\ell}_{\mathcal{S}}(h)$ as 
\begin{equation} 
\hat{\mathcal{L}}^{\ell}_{\mathcal{S}}(h) = \frac{1}{n}\sum_{i=1}^n\text{KL}[P^*(F_Y|X =x_i, Y=y_i)||Q(F_Y|X=x_i)].
\end{equation}
To keep consistent with the definition of variational inference, we reformulate $\hat{\mathcal{L}}^{\ell}_{\mathcal{S}}(h)$ as 
\begin{equation} 
\hat{\mathcal{L}}^{\ell}_{\mathcal{S}}(h) = \frac{1}{n}\sum_{i=1}^n\{\text{KL}[Q(F_Y|X=x_i)||P^*(F_Y|X=x_i, Y=y_i)] + d\}.
\end{equation}
where $d = \text{KL}[P^*(F_Y|X, Y)||Q(F_Y|X)] - \text{KL}[Q(F_Y|X)||P^*(F_Y|X, Y)]$. 

Though KL-divergence is asymmetric, i.e., $\text{KL}[P||Q] \neq \text{KL}[Q||P]$, we show that $d$ converges to zero during minimizing $\hat{\mathcal{L}}^{\ell}_{\mathcal{S}}(h)$, because $P^*(F_Y|X, Y)$ is fixed and $Q(F_Y|X)$ is the only variable.
The proof for $d \rightarrow 0$ is presented in Appendix \ref{KL_asymmetric}.
Moreover, since $\mathcal{S}$ consists of \textit{i.i.d.}\ samples and KL-divergence is additive for independent distributions, we can finally derive
\begin{equation} 
\label{mlp_vi}
\underset{h \in \mathcal{H}}{\text{min}}n\hat{\mathcal{L}}^{\ell}_{\mathcal{S}}(h) = \underset{Q(F_Y|X)}{\text{min}}\text{KL}[Q_{\mathcal{S}_x}(F_Y|X)||P^*_{\mathcal{S}}(F_Y|X, Y)].
\end{equation}
As a result, minimizing $n\hat{\mathcal{L}}^{\ell}_{\mathcal{S}}(h)$ with the cross entropy loss is equivalent to Bayesian variational inference. 
In summary, Theorem 3 provides a novel probabilistic explanation for MLPs from the perspective of Bayesian variational inference. 
Specifically, the MLP formulates a family of Gibbs distribution $Q(F_Y|X)$, and minimizing $\hat{\mathcal{L}}^{\ell}_{\mathcal{S}}(h)$ aims to search an optimal distribution $Q(F_Y|X)$, which is closest to the truly posterior distribution $P^*(F_Y|X, Y)$ given $\mathcal{S}$.

\section{PAC-Bayesian Bounds for Multilayer Perceptrons}

In this section, we prove that MLPs with the cross entropy loss inherently guarantee PAC-Bayesian generalization bounds 
based on the ELBO derived from Bayesian variational inference.
Moreover, we derive a novel PAC-Bayesian generalization bound for MLPs.

\paragraph{Theorem 4:} given the samples $\mathcal{S}$ and the $\text{MLP} = \{{x; f_1;f_2; f_Y}\}$, if $\hat{\mathcal{L}}^{\ell}_{\mathcal{S}}(h)$ is the cross entropy, then minimizing $\hat{\mathcal{L}}^{\ell}_{\mathcal{S}}(h)$ inherently guarantees PAC-Bayesian generalization bounds for the MLP.
\begin{equation} 
\underset{h \in \mathcal{H}}{\text{min}}\hat{\mathcal{L}}^{\ell}_{\mathcal{S}}(h) \Rightarrow \underbrace{\underset{Q(F|X)}{\text{min}}\text{KL}[Q_{\mathcal{S}_x}(F_Y|X)||P^*_{\mathcal{S}}(F_Y|X, Y)]}_{\text{Bayesian variational inference}} \Rightarrow \underbrace{\underset{\rho}{\text{min }}n \underset{h \sim \rho}{\boldsymbol{E}}\hat{\mathcal{L}}^{\ell}_{\mathcal{S}}(h)+\text{KL}[\rho||\pi]}_{\text{PAC-Bayesian optimization}}.
\end{equation}

\textit{Proof:} we only need prove Bayesian variational inference inducing PAC-Bayesian generalization bounds based on Theorem 3.
Given the Bayesian variational inference (Equation \ref{mlp_vi}), we have 
\begin{equation}
\begin{split}
\text{KL}[{\scriptstyle Q_{\mathcal{S}_x}(F_Y|X)}||{\scriptstyle P^*_{\mathcal{S}}(F_Y|X, Y)}] = &\underset{Q(F_Y|X)}{\boldsymbol{E}}\text{log}{\scriptstyle Q_{\mathcal{S}_x}(F_Y|X)} -  \underset{Q(F_Y|X)}{\boldsymbol{E}} \text{log}{\scriptstyle P^*_{\mathcal{S}}(F_Y|X, Y)} \\
=& \underset{Q(F_Y|X)}{\boldsymbol{E}} \text{log}{\scriptstyle Q_{\mathcal{S}_x}(F_Y|X)} - \underset{Q(F_Y|X)}{\boldsymbol{E}} [\text{log}{\scriptstyle P^*_{\mathcal{S}}(F_Y,Y|X)} - \text{log}{\scriptstyle P_{\mathcal{S}}(Y|X)}].
\end{split}
\end{equation}
Since $\text{log}P_{\mathcal{S}}(Y|X)$ is constant with respect to $Q(F_Y|X)$, we can derive the ELBO as
\begin{equation}
\label{elbo}
\text{log}P_{\mathcal{S}}(Y|X) = \underbrace{\underset{Q(F_Y|X)}{\boldsymbol{E}}\text{log} {\scriptstyle P_{\mathcal{S}}(Y|F_Y,X)}-\text{KL}[{\scriptstyle Q_{\mathcal{S}_x}(F_Y|X)}|| {\scriptstyle P^*_{\mathcal{S}_x}(F_Y|X)}]}_{\text{ELBO}} + \text{KL}[{\scriptstyle Q_{\mathcal{S}_x}(F_Y|X)}||{\scriptstyle P^*_{\mathcal{S}}(F_Y|X, Y)}],
\end{equation}
where $P_{\mathcal{S}}(Y|X) = \prod_{i=1}^nP(Y=y_i|X=x_i)$, $P_{\mathcal{S}}(Y|F_Y,X) = \prod_{i=1}^nP(Y=y_i|F_Y, X=x_i)$, $Q_{\mathcal{S}_x}(F_Y|X) = \prod_{i=1}^nQ(F_Y|X=x_i)$, and $P^*_{\mathcal{S}_x}(F_Y|X) = \prod_{i=1}^nP^*(F_Y|X=x_i)$.
The detailed derivation of the ELBO is presented in Appendix \ref{elbo_app}.

\pagebreak
Since $P^*(F_Y|X)$ is the truly prior distribution of the true hypothesis $h^*$, we have $\pi := P^*_{\mathcal{S}_x}(F_Y|X)$. 
Since $Q(F_Y|X)$ infers the truly posterior distribution $P^*(F_Y|X, Y)$ of $h$, we have $\rho := Q_{\mathcal{S}_x}(F_Y|X)$.
Corollary 1 induces $n\hat{\mathcal{L}}^{\ell}_{\mathcal{S}}(h) = -\text{log}P_{\mathcal{S}}(Y|F_Y,X)$ because we can express $P(Y=y_i|F_Y,X=x_i)$ as $Q(F_Y = y_i|X=x_i)$. 
As a result, we derive
\begin{equation}
\label{elbo2} 
\text{ELBO} = -(n\underset{Q(F_Y|X)}{\boldsymbol{E}}{ \hat{\mathcal{L}}^{\ell}_{\mathcal{S}}(h)}+\text{KL}[{ Q_{\mathcal{S}_x}(F_Y|X)}|| { P^*_{\mathcal{S}_x}(F_Y|X)}]).
\end{equation} 
It implies that maximizing ELBO is equivalent to minimizing PAC-Bayesian generalization bounds, thus minimizing $\hat{\mathcal{L}}^{\ell}_{\mathcal{S}}(h)$ inherently guarantees PAC-Bayesian generalization bounds for the MLP.

Theorem 4 theoretically explains why MLPs can achieve great generalization performance without explicit regularization.
More specifically, the ELBO indicates that minimizing $\hat{\mathcal{L}}^{\ell}_{\mathcal{S}}(h)$ makes MLPs not only to learn the likelihood $P_{\mathcal{S}}(Y|F_Y,X)$ but also to learn the prior knowledge via minimizing $\text{KL}[{\scriptstyle Q_{\mathcal{S}_x}(F_Y|X)}|| {\scriptstyle P^*_{\mathcal{S}_x}(F_Y|X)}]$.
In other words, minimizing $\hat{\mathcal{L}}^{\ell}_{\mathcal{S}}(h)$ inherently guarantees PAC-Bayesian generalization bounds for MLPs, even though they do not incorporate explicit regularization.

\paragraph{Corollary 2:} PAC-Bayesian generalization bounds for the MLP.

Given a distribution $\mathcal{D}$ over $\mathcal{X} \times \mathcal{Y}$, the samples $\mathcal{S}$, a probabilistic hypothesis set $\mathcal{H}$ derived from the $\text{MLP} = \{{x; f_1;f_2; f_Y}\}$, the cross entropy loss $\ell: \mathcal{F}\times \mathcal{X} \times \mathcal{Y} \rightarrow \mathbb{R}^+$ \footnote{The previous works prove PAC-Bayesian bounds being valid for unbounded loss functions \cite{germain2016pac, alquier2016properties}.}, a real number $b > 0$, and a real number $\delta \in (0,1]$, with probability at least $1-\delta$ over the samples $\mathcal{S} \sim \mathcal{D}^n$, we have
\begin{equation} 
\label{bound}
\forall Q(F_Y|X) \text{ on } \mathcal{H}: \underset{{h \sim Q(F_Y|X)}}{\boldsymbol{E}}\mathcal{L}^{\ell}_{\mathcal{D}}(h) \leq \frac{b}{1-e^{-b}}\{1-\frac{\sqrt[n]{\delta\cdot P_{\mathcal{S}}(Y|X)}}{\text{exp}(\frac{1}{n}\sum_{i=1}^n[E_{yy_i}(x_i)+\text{log}Z_{\text{MLP}}(x_i)])}\}.
\end{equation}
where $P_{\mathcal{S}}(Y|X) = \prod_{i=1}^nP_{Y|X}(y_i|x_i)$, $E_{yy_i}(x_i) = -f_{yy_i}(f_2(f_1(x_i)))$ is the energy of $y_i$ given $x_i$, and $Z_{\text{MLP}}(x_i) = \sum_{l=1}^L \sum_{k=1}^K \sum_{t=1}^T P(F_Y, F_2, F_1|X=x_i)$ is the partition function.

\textit{Proof:} the result is derived by two steps: (i) substituting $n \underset{h \sim \rho}{\boldsymbol{E}}\hat{\mathcal{L}}^{\ell}_{\mathcal{S}}(h)+\text{KL}[\rho||\pi]$ in Equation \ref{pac_bayesian} for the $\text{ELBO} = \text{log}P_{\mathcal{S}}(Y|X) - \text{KL}[{\scriptstyle Q_{\mathcal{S}_x}(F_Y|X)}||{\scriptstyle P^*_{\mathcal{S}}(F_Y|X, Y)}]$ in Equation \ref{elbo}; and (ii) substituting $\text{KL}[{\scriptstyle Q_{\mathcal{S}_x}(F_Y|X)}||{\scriptstyle P^*_{\mathcal{S}}(F_Y|X, Y)}]$ for $\sum_{i=1}^n[E_{yy_i}(x_i) + \text{log}Z_{\text{MLP}}(x_i)]$ based on Equation \ref{mlp_vi} and \ref{energy_minimization}.

Corollary 1 implies that when minimizing $\hat{\mathcal{L}}^{\ell}_{\mathcal{S}}(h)$ to zero, $\frac{1}{n}\sum_{i=1}^n[E_{yy_i}(x_i)+\text{log}Z_{\text{MLP}}(x_i)] \rightarrow 0$ and the proposed PAC-Bayesian generalization bound would be entirely determined by the marginal likelihood $P_{\mathcal{S}}(Y|X)$, which is consistent with the theoretical connection between PAC-Bayesian theory and Bayesian inference proposed by Germain \textit{et al.} \cite{germain2016pac}. 

Based on Corollary 2, we derive an applicable bound for examining the generalization of MLPs. 
Though $P(Y|X)$ is unknown, the $\textit{i.i.d.}$ assumption for $\mathcal{S}$ induces $\sqrt[n]{ P_{\mathcal{S}}(Y|X)} = P(Y|X)$ being a constant as long as the number of different labels are equivalent.
As a result, we have
\begin{equation}
\text{the PAC-Bayesian generalization bound} \propto \text{exp}(\frac{1}{n}\sum_{i=1}^n[E_{yy_i}(x_i)+\text{log}Z_{\text{MLP}}(x_i)]).
\end{equation} 

Recall that $Z_{\text{MLP}}(x_i) = \sum_{l=1}^L \sum_{k=1}^K \sum_{t=1}^T Q(F_Y, F_2, F_1|X=x_i)$ sums up all the cases of $F_Y$, i.e., $Z_{\text{MLP}}(x_i)$ is a constant with respect to $Q(F_Y|X)$.
In addition, we show that the functionality of $Z_{\text{MLP}}(x_i)$ is only to guarantee the validity of $Q(F_Y|X)$ (Appendix \ref{energy_log_partition}).
Therefore, $\sum_{i=1}^n\text{log}Z_{\text{MLP}}(x_i)$ can be viewed as a constant for deriving PAC-Bayesian generalization bounds for MLPs, thus the bound can be further relaxed as 
\begin{equation}
\text{the PAC-Bayesian generalization bound} \propto \frac{1}{n}\sum_{i=1}^nE_{yy_i}(x_i).
\end{equation} 
We use the above formula to approximate the generalization bound for subsequent experiments.

\section{Experimental Results}
\label{simulations}

In this section, we demonstrate the proposed PAC-Bayesian generalization bound on MLPs in four different cases: (i) different training algorithms, (ii) different architectures of MLPs, (iii) different sample sizes, and (iv) different labels, namely real labels and random labels.

All the experimental results are derived based on the $\text{MLP} = \{{x; f_1;f_2; f_Y}\}$ classifying the MNIST dataset \cite{lecun_cnns}.
Since the dimension of a single MNIST image is $28 \times 28$, $x$ has $M = 784$ input nodes.
Since the MNIST dataset consists of $10$ different labels, ${f_Y}$ has $L = 10$ output nodes.
All the activation functions are ReLU $\sigma(z):=\text{max}(0,z)$.
Because of space limitation, we present more experimental results based on MLPs on other benchmark datasets in Appendix \ref{mlp_fashin_mnist}.

\textbf{The bound can reflect the generalization of MLPs with different number of neurons \cite{generalization_regularization1}.}

\begin{figure}[!t]
\centering
\includegraphics[scale=0.5]{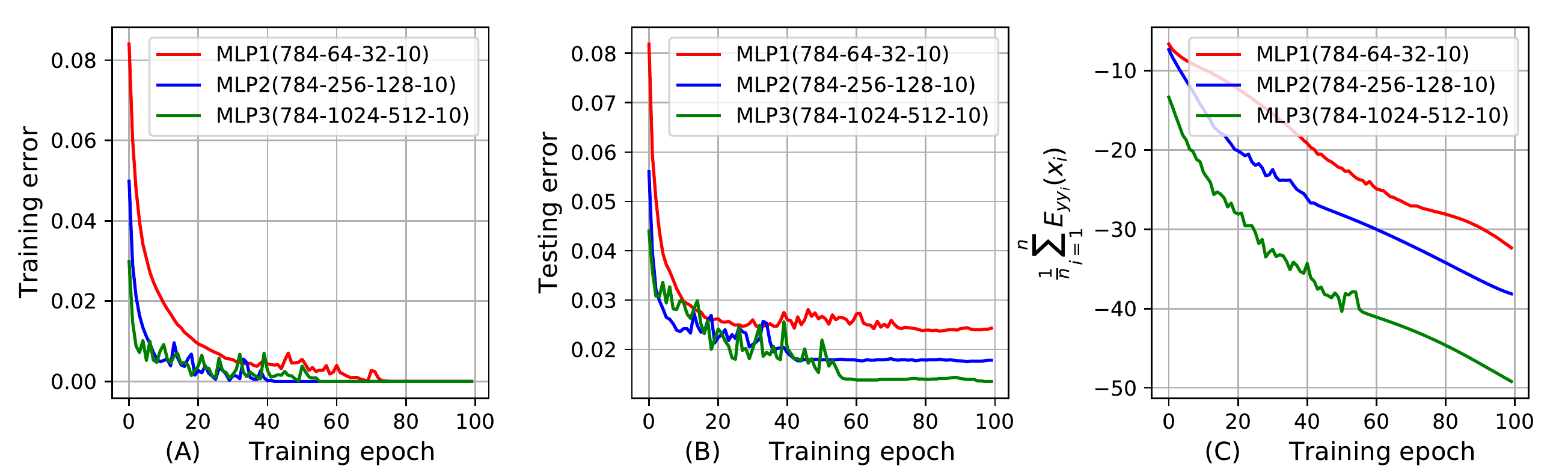}
\caption{
(A), (B), and (C) show the training error, the testing error, and the bound over 100 training epochs, respectively.
In the legend, (M-T-K-L) denotes the number of nodes/neurons in each layer, e.g., MLP1 has $T = 64$ neurons in $f_1$ and $K = 32$ neurons in $f_2$.
}
\label{Img_generalization_architecture}
\end{figure}

We design three MLPs (abbr. MLP1, MLP2, MLP3). 
They have the same architecture except the number of neurons in hidden layers, which are summarized in Figure \ref{Img_generalization_architecture}.
We use the same SGD optimization algorithm to train the three MLPs and all the three MLPs achieve zero training error.
We visualize the training error, the testing error, and the bound in Figure \ref{Img_generalization_architecture}, which shows that the bound has the same architecture dependence as the testing error.
In other words, the MLP with the more number of neurons has the lower testing error and bound.

\textbf{The bound can reflect the effect of SGD optimization on the generalization of MLPs \cite{generalization_regularization1}.}

\begin{figure}[!b]
\centering
\includegraphics[scale=0.52]{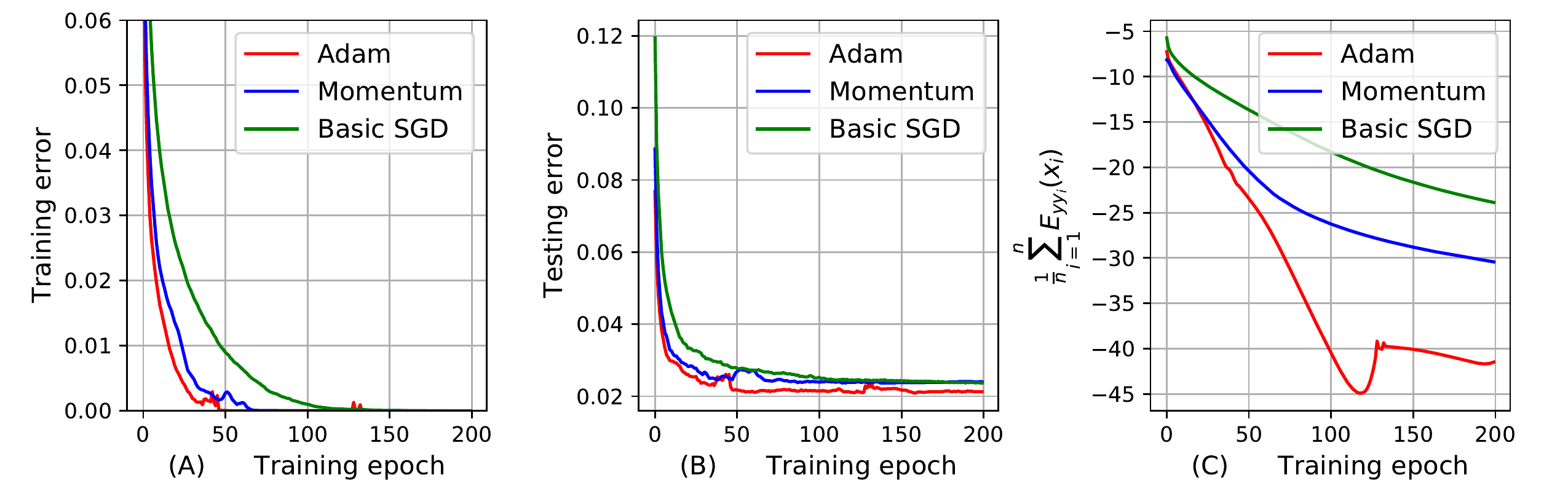}
\caption{
(A), (B), and (C) show the training error, the testing error, and the proposed PAC-Bayesian generalization bound over 200 training epochs, respectively. 
}
\label{Img_generalization_SGDs}
\end{figure}

We use three different training algorithms (the basic SGD \cite{backpropagation}, Momentum \cite{qian1999momentum}, and Adam \cite{kingma2014adam}) to train MLP2 200 epochs, and visualize the training error, the testing error, and the generalization bound in Figure \ref{Img_generalization_SGDs}.
We can observe that the bounds of MLP2 with different training algorithms is consistent with the corresponding testing errors of MLP2.
For example, MLP2 with Adam training algorithm achieves the smallest testing error, thus the bound of MLP2 with Adam is also smaller than Momentum and the basic SGD.

\begin{figure}
\centering
\includegraphics[scale=0.45]{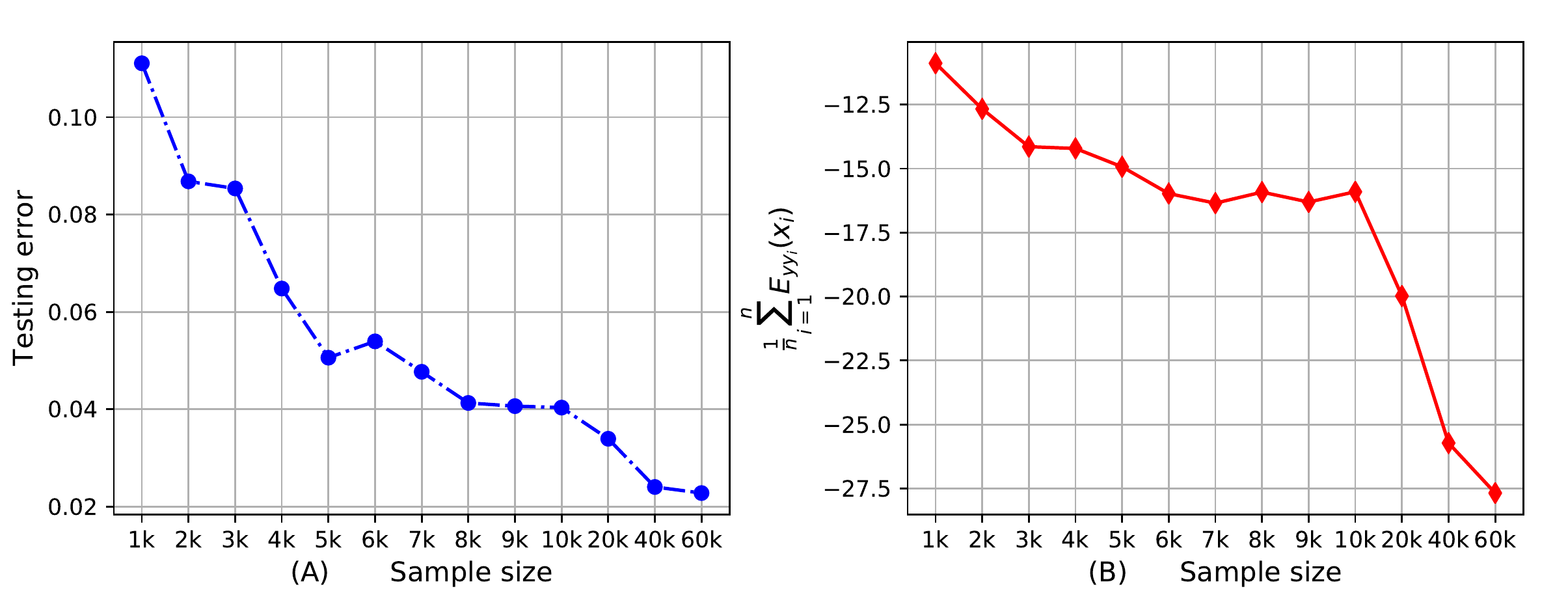}
\caption{
(A) and (B) show the testing error and the bound given different training sample sizes. 
}
\label{Img_generalization_sample_size}
\end{figure}

\textbf{The bound can reflect the same sample size dependence as the testing error \cite{nagarajan2019uniform}.}

We create 13 different subsets of the MNIST dataset with different number of training samples, and train MLP2 to classify these samples until the training error becomes zero.
Figure \ref{Img_generalization_sample_size} shows the testing error and the bound with different sample sizes.
The bound shows a general decreasing trend as the training sample size increases, which is consistent with the variation of the testing error.
  

\textbf{The bound can reflect the generalization performance of MLPs given random labels.}

\begin{figure}[htp]
\centering
\includegraphics[scale=0.45]{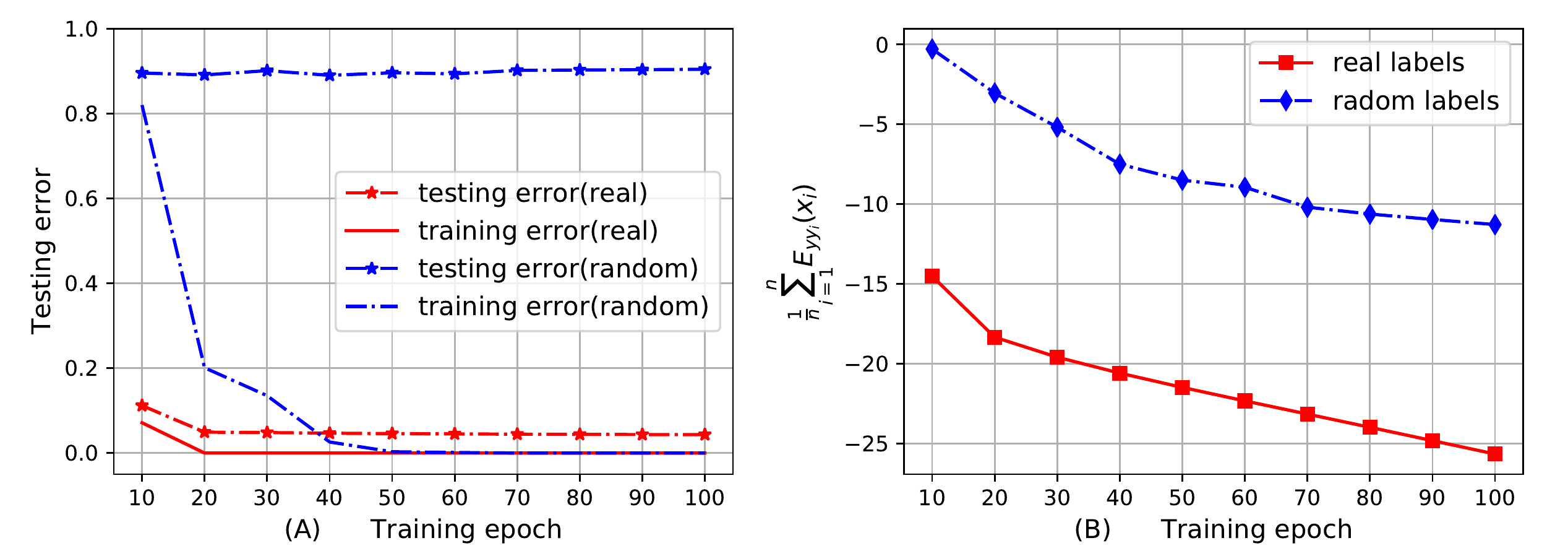}
\caption{
(A) and (B) show the training/testing error and the bound given real and random labels. 
}
\label{Img_generalization_random}
\end{figure}

Zhang \textit{et al.}\cite{generalization_regularization} show that DNNs achieve low testing error when they are trained by random labels. 
We train MLP3 100 epochs to classify 10,000 samples of the MNIST dataset with random labels, and visualize the training error, the testing error and the bound in Figure \ref{Img_generalization_random}.

We observe that the bound of MLP3 trained by random labels is higher than real labels, which is consistent with the testing error based on random labels is higher than real labels.
In particular, we can theoretically explain why MLP3 trained by random labels has the higher testing error.
Random labels imply that $Y$ not depends on $F_Y$ and $X$, i.e., $P_{\mathcal{S}}(Y|X) = P_{\mathcal{S}}(Y|F_Y,X) = P_{\mathcal{S}}(Y)$, thus the ELBO (Equation \ref{elbo}) can be reformulated as
\begin{equation}
\text{log}P_{\mathcal{S}}(Y) = \underbrace{\text{log} { P_{\mathcal{S}}(Y)}-\text{KL}[{\scriptstyle Q_{\mathcal{S}_x}(F_Y|X)}|| {\scriptstyle P^*_{\mathcal{S}_x}(F_Y|X)}]}_{\text{ELBO}} + \text{KL}[{\scriptstyle Q_{\mathcal{S}_x}(F_Y|X)}||{\scriptstyle P^*_{\mathcal{S}}(F_Y|X, Y)}].
\end{equation}
As a result, minimizing $\text{KL}[{\scriptstyle Q_{\mathcal{S}_x}(F_Y|X)}||{\scriptstyle P^*_{\mathcal{S}}(F_Y|X, Y)}]$ equals minimizing $\text{KL}[{\scriptstyle Q_{\mathcal{S}_x}(F_Y|X)}|| {\scriptstyle P^*_{\mathcal{S}_x}(F_Y|X)}]$, which is only one part of the PAC-Bayesian optimization (Equation \ref{pac_equivalence}).
Therefore, MLPs with random labels cannot guarantee PAC-Bayesian generalization bounds.

\section{Conclusions}

In this work, we prove that MLPs with the cross entropy loss inherently guarantee PAC-Bayesian generalization bounds, and minimizing PAC-Bayesian generalization bounds for MLPs is equivalent to maximizing the ELBO. 
In addition, we derive a novel PAC-Bayesian generalization bound correctly explaining the generalization performance of MLPs.
Extending the proposed PAC-Bayesian generalization bounds to general DNNs is promising for future work.

\clearpage
\small
\bibliography{neurips_2020}

\clearpage
\appendix
\section*{Appendix}
\renewcommand{\thesubsection}{\Alph{subsection}}

\subsection{Related works about PAC-Bayesian generalization bounds on DNNs}
\label{related_works}

In this appendix, we briefly discuss previous works using PAC-Bayesian theory to derive the generalization bound for neural networks, and explain what is the difference to the current paper. 

To derive generalization bounds for neural networks based on PAC-Bayesian theory, it is prerequisite that obtaining the KL-divergence between the posterior distribution and the prior distribution of the hypothesis expressed by neural networks. 

\textbf{Behnam Neyshabur \textit{et al.} \cite{neyshabur2017pac}.} This paper combines the perturbation bound and PAC-Bayesian theory to derive the generalization bound of neural networks. The authors approximate the KL-divergence involved PAC-Bayesian bounds by the perturbation bound based on the assumption that the prior distribution is Gaussian and the perturbation is Gaussian as well. 

\textbf{Wenda Zhou \textit{et al.} \cite{zhou2018non}.} In the context of compressing neural networks, this paper propose a generalization bound of neural networks through estimating the KL-divergence involved PAC-Bayesian bounds by the number of bits required to describe the hypothesis.

\textbf{Gintare Karolina Dziugaite \textit{et al.} \cite{dziugaite2017computing}.} To derive PAC-Bayesian generalization bounds for stochastic neural networks, this paper restricts the prior distribution and the posterior distribution as a family of multivariate normal distributions for simplifying the KL-divergence involved PAC-Bayesian bounds.

\textbf{Letarte Ga{\"e}l \textit{et al.} \cite{letarte2019dichotomize}.} This paper studies PAC-Bayesian generalization bounds for binary activated deep neural networks, which can be explained as a specific linear classifier, and the authors assume the prior distribution and the posterior distribution being Gaussian. As a result, PAC-Bayesian generalization bounds are determined by the summation of the Gaussian error function and the $\ell_2$ norm of the weights.

We can find that most existing works use relaxation or approximation to estimate the KL-divergence involved PAC-Bayesian bounds.
However, the relaxation or approximation not only introduce estimation error of the generalization bound for neural networks but also make difficult to clearly explain why neural networks can achieve impressive generalization performance without explicit regularization.
The limitation of existing PAC-Bayesian generalization bounds triggers a fundamental question:

\hfill \break
\centerline{\textbf{Can we establish an explicit probabilistic representation for DNNs and}}
\centerline{\textbf{derive an explicit  PAC-Bayesian generalization bound?}}
\hfill \break
The current paper attempts to derive PAC-Bayesian generalization bounds for MLPs based on explicit probabilistic representations rather than approximation or relaxation.
It is mainly inspired by the previous work \cite{germain2016pac}, in which Germain \textit{et al.} bridge Bayesian inference and PAC-Bayesian theory.
They prove that minimizing the PAC-Bayesian generalization bound is equivalent to maximizing the Bayesian marginal likelihood if the loss function is the negative log-likelihood of the hypothesis. 
However, the paper only focuses on linear regression and model selection problem, and does not discuss how to use Bayesian inference to explain neural networks

The current paper is also inspired by the previous works \cite{mandt2017stochastic, chaudhari2018stochastic, Boltzmann_machine, lin2017does}.
Mandt \textit{et al.} provide novel explanations for SGD from the viewpoint of Bayesian inference, especially Bayesian variational inference \cite{mandt2017stochastic, chaudhari2018stochastic}.
Mehta \textit{et al.} demonstrate that the distribution of a hidden layer can be formulated as a Gibbs distribution in the Restricted Boltzmann Machine (RBM) \cite{Boltzmann_machine, lin2017does}.

\subsection{The probability space $(\Omega, \mathcal{T}, Q_F)$ for a fully connected layer}
\label{prob_space}

\begin{figure}
\centering
\includegraphics[scale=0.7]{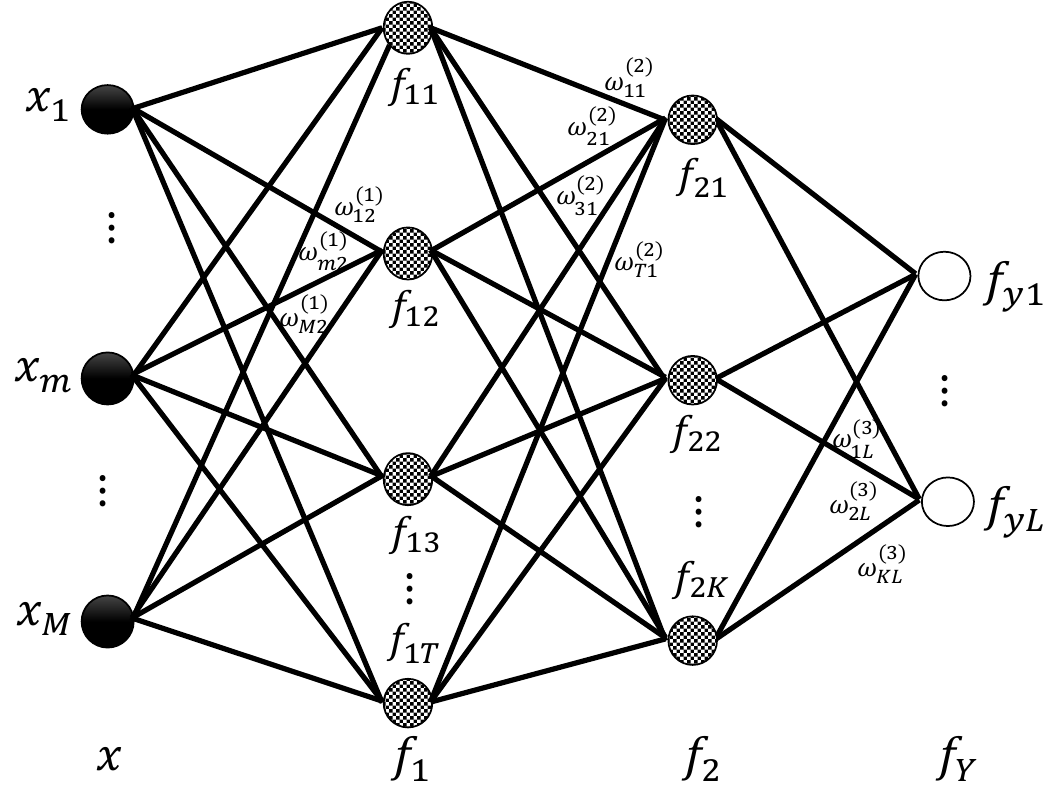}
\caption{
The $\text{MLP} = \{{x; f_1; f_2; f_Y}\}$. The input layer ${x}$ has $M$ nodes, and ${f_1}$ has $T$ neurons, i.e., ${f_1} = \{f_{1t} = \sigma_1[g_{1t}({x})]\}_{t=1}^{T}$, where $g_{1t}(x) = \sum_{m=1}^M\omega^{(1)}_{mt} \cdot x_{m} + b_{1t}$ is the $t$th linear function with $\omega^{(1)}_{mt}$ being the weight of the edge between $x_m$ and $f_{1t}$ and $b_{1t}$ being the bias, and $\sigma_1(\cdot)$ is a non-linear activation function, e.g., ReLU function.
Similarly, ${f_{2}} = \{f_{2k}=\sigma_2[g_{2k}({f_1})]\}_{k=1}^K$ has $K$ neurons, where $g_{2k}({f_1}) = \sum_{t=1}^T\omega^{(2)}_{tk} \cdot f_{1t} + b_{2k}$.
The output layer ${f_Y}$ is the softmax, thus $f_{yl} = \frac{1}{{Z(f_2)}}\text{exp}[g_{yl}(f_2)]$, where $g_{yl}(f_2)=\sum_{k=1}^K\omega^{(3)}_{kl} \cdot f_{2k} + b_{yl}$ and $Z(f_2) = \sum_{l=1}^L\text{exp}[g_{yl}({f_2})]$.
}
\label{fig_mlp_app}
\end{figure}

\subsubsection{The proof of the probability space definition $(\Omega, \mathcal{T}, Q_F)$}

We use the mathematical induction to prove the probability space $(\Omega, \mathcal{T}, Q_F)$ for all the fully connected layers of the MLP in Figure \ref{fig_mlp_app}.
Above all, we define three probability space, i.e., $(\Omega_1, \mathcal{T}, Q_{F_1})$, $(\Omega_2, \mathcal{T}, Q_{F_2})$, and $(\Omega_Y, \mathcal{T}, Q_{F_Y})$, for the three layers of the MLP, i.e., ${f_1}$, ${f_2}$, and ${f_Y}$, respectively. 
Subsequently, we first demonstrate that $Q_{{F_Y}}$ is a Gibbs distribution, and then we prove that $Q_{{F_2}}$ and $Q_{{F_1}}$ are Gibbs distributions based on $Q_{{F_Y}}$ and $Q_{{F_2}}$ being Gibbs distributions, respectively.

Since the output layer ${f_Y}$ is the softmax, each output node $f_{yl}$ can be formulated as 
\begin{equation}
\label{Gibbs_fY}
\begin{split}
f_{yl} &= \frac{1}{Z_{F_Y}}\text{exp}[\sum_{k=1}^K\omega^{(3)}_{kl} \cdot f_{2k} + b_{yl}]\\
&= \frac{1}{Z_{F_Y}}\text{exp}[\langle \boldsymbol{\omega}^{(3)}_l, {f_2} \rangle + b_{yl}],\\
\end{split}
\end{equation}
where $Z_{F_Y} = \sum_{l=1}^L\text{exp}(f_{yl})$ is the partition function and $\boldsymbol{\omega}^{(3)}_{l} = \{\omega^{(3)}_{kl}\}_{k=1}^K$.

Comparing Equation \ref{Gibbs_f} and \ref{Gibbs_fY},  we can derive that ${f_Y}$ forms a Gibbs distribution $Q_{F_Y}(\boldsymbol{\omega}^{(3)}_l) = f_{yl}$ to measure the probability of $\boldsymbol{\omega}^{(3)}_l$ occurring in ${f_2}$, which is consistent with the definition of $(\Omega_Y, \mathcal{T}, Q_{F_Y})$.

Based on the properties of the exponential function, namely ${\textstyle \text{exp}(a+b) = \text{exp}(a)\cdot \text{exp}(b)}$ and ${\textstyle \text{exp}(a\cdot b) = [\text{exp}(b)]^{a}}$, we can reformulate $Q_{{F_Y}}(\boldsymbol{\omega}^{(3)}_l)$ as 
\begin{equation} 
Q_{{F_Y}}(\boldsymbol{\omega}^{(3)}_l) = \frac {1}{Z'_{F_Y}}\prod_{k=1}^K[\text{exp}(f_{2k})]^{\omega^{(3)}_{kl}},
\end{equation}
where $Z'_{F_Y} = Z_{F_Y}/\text{exp}(b_{yl})$. 
Since $\{\text{exp}(f_{2k})\}_{k=1}^K$ are scalar, we can introduce a new partition function ${\textstyle Z_{{F_2}} = \sum_{k=1}^K\text{exp}(f_{2k})}$ such that $\{\frac{1}{Z_{{F_2}}}\text{exp}(f_{2k})\}_{k=1}^K$ becomes a probability measure, thus we can reformulate $Q_{{F_Y}}(\boldsymbol{\omega}^{(3)}_l) $ as a Product of Expert (PoE) model \cite{CD}
\begin{equation} 
\label{FoE_fY}
Q_{{F_Y}}(\boldsymbol{\omega}^{(3)}_l) =\frac {1}{Z''_{F_Y}}\prod_{k=1}^K[\frac{1}{Z_{{F_2}}}\text{exp}(f_{2k})]^{\omega^{(3)}_{kl}},
\end{equation}
where ${\textstyle Z''_{F_Y} = Z_{F_Y}/[\text{exp}(b_{yl}) \cdot \prod_{k=1}^K[Z_{{F_2}}]^{\omega^{(3)}_{kl}}}]$, especially each expert is defined as $\frac{1}{Z_{{F_2}}}\text{exp}(f_{2k})$.

It is noteworthy that all the experts $\{\frac{1}{Z_{{F_2}}}\text{exp}(f_{2k})\}_{k=1}^K$ form a probability measure and establish an exact one-to-one correspondence to all the neurons in ${f_2}$, thus the distribution of ${f_2}$ can be expressed as
\begin{equation}
\label{Gibbs_f2}
Q_{{F_2}} = \{\frac {1}{Z_{{F_2}}}\text{exp}(f_{2k})\}_{k=1}^K.
\end{equation}

Since $\{f_{2k} = \sigma_2(\sum_{t=1}^T\omega^{(2)}_{tk} \cdot f_{1t} + b_{2k})\}_{k=1}^K$, $Q_{{F_2}}$ can be extended as 
\begin{equation}
\begin{split}
Q_{{F_2}}(\boldsymbol{\omega}^{(2)}_{k}) &= \frac {1}{Z_{{F_2}}}\text{exp}[\sigma_2(\sum_{t=1}^T\omega^{(2)}_{tk} \cdot f_{1t} + b_{2k})]\\
&= \frac {1}{Z_{{F_2}}}\text{exp}[\sigma_2(\langle \boldsymbol{\omega}^{(2)}_k, {f_1} \rangle + b_{2k})],
\end{split}
\end{equation}
where $Z_{{F_2}} = \sum_{k=1}^K\text{exp}(f_{2k})$ is the partition function and $\boldsymbol{\omega}^{(2)}_{k} = \{\omega^{(2)}_{tk}\}_{t=1}^T$.
We can conclude that ${f_2}$ forms a Gibbs distribution $Q_{{F_2}}(\boldsymbol{\omega}^{(2)}_k)$ to measure the probability of $\boldsymbol{\omega}^{(2)}_k$ occurring in ${f_1}$, which is consistent with the definition of $(\Omega_2, \mathcal{T}, Q_{{F_2}})$.

Due to the non-linearity of $\sigma_2(\cdot)$, we cannot derive $Q_{{F_2}}(\boldsymbol{\omega}^{(2)}_{k})$ being a PoE model only based on the properties of exponential functions.
Alternatively, the equivalence between the Stochastic Gradient Descent (SGD) and the first order approximation \cite{GD_Taylor} indicates that $\sigma_2(\sum_{t=1}^T\omega^{(2)}_{tk} \cdot f_{1t} + b_{2k})]$ can be approximated as 
\begin{equation} 
\sigma_2(\sum_{t=1}^T\omega^{(2)}_{tk} \cdot f_{1t} + b_{2k})] \approx C_{21} \cdot [\sum_{t=1}^T\omega^{(2)}_{tk} \cdot f_{1t} + b_{2k}]  + C_{22},
\end{equation}
where $C_{21}$ and $C_{22}$ only depend on the activations $\{f_{1t}\}_{t=1}^T$ in the previous training iteration, thus they can be regarded as constants and absorbed by $\omega^{(2)}_{tk}$ and $b_{2k}$.
The proof for the approximation in Appendix \ref{GD_Approx}.

Therefore, $Q_{F_2}$ still can be modeled as a PoE model
\begin{equation} 
\label{FoE_f2}
Q_{{F_2}} = \{\frac {1}{Z''_{{F_2}}}\prod_{t=1}^T[\frac{1}{Z_{{F_1}}}\text{exp}(f_{1t})]^{\omega^{(2)}_{tk}}\}_{k=1}^K,
\end{equation}
where ${Z''_{{F_2}} = Z_{{F_2}}/[\text{exp}(b_{2k})\cdot \prod_{t=1}^TZ_{{F_1}}^{\omega^{(2)}_{tk}}}]$ and the partition function is ${ Z_{{F_1}} = \sum_{t=1}^T\text{exp}(f_{1t})}$.

Similar to $Q_{{F_2}}(\boldsymbol{\omega}^{(2)}_{k})$, we can derive the probability measure of ${f_1}$ as
\begin{equation} 
\label{Gibbs_f1}
\begin{split}
Q_{{F_1}}(\boldsymbol{\omega}^{(1)}_{t}) &= \frac{1}{Z_{{F_1}}}\text{exp}(f_{1t}) = \frac{1}{Z_{{F_1}}}\text{exp}[\sigma_1(\sum_{m=1}^M \omega^{(1)}_{mt} \cdot x_m + b_{1t})]\\
&= \frac{1}{Z_{{F_1}}}\text{exp}[\sigma_1(\langle \boldsymbol{\omega}^{(1)}_{t}, \boldsymbol{x} \rangle + b_{1t})],\\
\end{split}
\end{equation}
where $Z_{F_1} = \sum_{t=1}^N\text{exp}(f_{1t})$ is the partition function and $\boldsymbol{\omega}^{(1)}_{t} = \{\omega^{(1)}_{mt}\}_{m=1}^M$.
We can conclude that ${f_1}$ forms a Gibbs distribution $Q_{{F_1}}(\boldsymbol{\omega}^{(1)}_t)$ to measure the probability of $\boldsymbol{\omega}^{(1)}_t$ occurring in ${x}$, which is consistent with the definition of $(\Omega_1, \mathcal{T}, Q_{F_1})$.
Overall, we prove that the proposed probability space is valid for all the fully connected layers in the $\text{MLP} = \{x, f_1, f_2, f_Y\}$. 
Also note that we can easily extend the probability space to an arbitrary fully connected layer through properly changing the script.

\subsubsection{The validation of the probability space definition $(\Omega, \mathcal{T}, Q_F)$}

\paragraph{Setup.}

\begin{figure}[!b]
\centering
\includegraphics[scale=0.7]{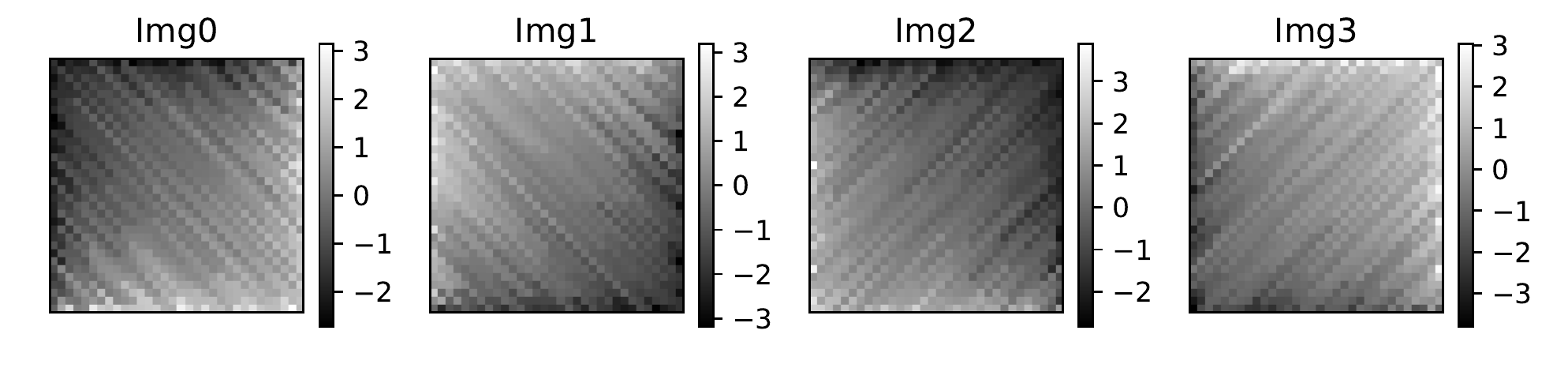}
\caption{\small{
Four synthetic images with different features. 
All images are sampled from $\mathcal{N}(0, 1)$. The only difference is that they are sorted by the ascending or descending order in two different diagonal directions.
Img0 and Img1 are sorted in the primary diagonal direction by the ascending order and the descending order, respectively.
Img2 and Img3 are sorted in the secondary diagonal direction by the ascending order and the descending order, respectively.
}}
\label{fig_iid_images1}
\end{figure}

We generate a synthetic dataset to validate the probability space for a fully connected layer.
The dataset consists of 256 $32 \times 32$ grayscale images and each image is sampled from the Gaussian distribution $\mathcal{N}(0, 1)$.
To guarantee that the synthetic dataset has certain features that can be learned by MLPs, we sort each image in the primary or secondary diagonal direction by the ascending or descending order.
As a result, the synthetic dataset has four features/classes, and the corresponding images are drawn in Figure \ref{fig_iid_images1}.

There is an important reason why we generate the synthetic dataset.
Since most benchmark dataset contains very complex features, we cannot explicitly explain the internal logic of MLPs.
As a comparison, the synthetic dataset only has four simple features.
Therefore, we can clearly demonstrate the proposed probability space for a fully connected layer by simply visualizing the weights of the layer.

To classify the synthetic dataset, we specify the MLP as follows: (i) since each synthetic image is $32 \times 32$, the input layer ${x}$ has $M = 1024$ nodes, (ii) two hidden layers ${f_1}$ and ${f_2}$ have $T = 8$ and $K = 6$ neurons, respectively, and (iii) the output layer ${f_Y}$ has $L = 4$ nodes. In addition, all the activation functions are chosen as ReLU $\sigma(z) = \text{max}(0, z)$ unless otherwise specified.

\paragraph{Validation.}

\begin{figure}[!t]
\centering
\includegraphics[scale=0.6]{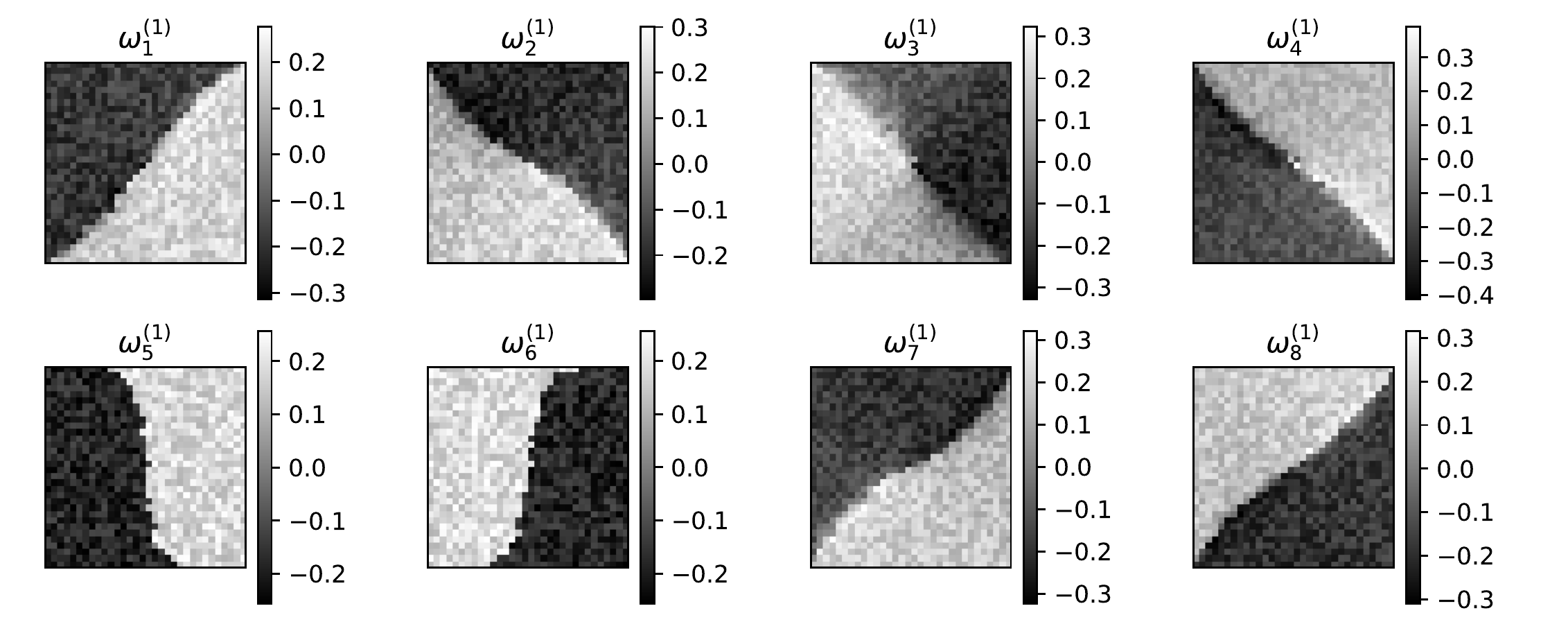}
\caption{\small{
The eight possible outcomes $\{\boldsymbol{\omega}^{(1)}_{t}\}_{t=1}^{8}$ represented by the learned weights of the eight neurons, where $\boldsymbol{\omega}^{(1)}_t = \{\omega^{(1)}_{mt}\}_{m=1}^{1024}$.
All the weights $\{\omega^{(1)}_{mt}\}_{m=1}^{1024}$ are reshaped to $32 \times 32$ dimension for visualizing the spatial structure.
}}
\label{Img_mlp_gibbs_sim1_features}
\end{figure}

To demonstrate $(\Omega, \mathcal{T}, Q_F)$ for all the fully connected layers in the $\text{MLP} = \{{x; f_1; f_2; f_Y}\}$, we only need to demonstrate $(\Omega_1, \mathcal{T}, Q_{F_1})$ for the first layer ${f_1}$, because we derive $(\Omega, \mathcal{T}, Q_F)$ for each layer in the back-forward direction based on the mathematical induction.

First, we demonstrate the sample space $\Omega_1 = \{\boldsymbol{\omega}^{(1)}_{t}\}_{t=1}^T$ for ${f_1}$.
More specifically, we use the synthetic dataset to train the MLP 100 epochs (the training accuracy is $100\%$) and visualize the learned weights of the eight neurons in ${f_1}$, i.e., $\boldsymbol{\omega}^{(1)}_{t} = \{\omega^{(1)}_{mt}\}_{m=1}^{1024}, t \in [1,8],$ in Figure \ref{Img_mlp_gibbs_sim1_features} .
We can find that $\boldsymbol{\omega}^{(1)}_{t}$ can be regarded as a feature of the input.
For example, $\boldsymbol{\omega}^{(1)}_{1}$ describes a feature indicating the difference along the secondary diagonal direction, because it has low magnitude at top-left positions and high magnitude at bottom-right positions.
In addition, $\boldsymbol{\omega}^{(1)}_{1}$ is strongly related to Img0 in Figure \ref{fig_iid_images1}, because the later has the same spatial feature.
In the context of probability space, we define an experiment as inputting Img0 into ${f_1}$, thus $\boldsymbol{\omega}^{(1)}_{1}$ can be viewed as a possible outcome of the experiment.
In addition, to keep consistent with the definition of outcome, we suppose $\boldsymbol{\omega}^{(1)}_{2}$ and $\boldsymbol{\omega}^{(1)}_{3}$ being mutually exclusive, because they have different weights at the same position, i.e., $\forall m, \omega^{(1)}_{m2} \neq \omega^{(1)}_{m3}$.

\begin{table}[!b]
\caption{The Gibbs probability of ${f_1}$ given the synthetic images in Figure \ref{fig_iid_images1}}
\label{fc_gibbs}
\vskip 0.1in
\begin{center}
\begin{small}
\begin{threeparttable}
\begin{tabular}{c|cccccccc}
\toprule
 & $\boldsymbol{\omega^{(1)}_{1}}$ & $\boldsymbol{\omega^{(1)}_{2}}$ & $\boldsymbol{\omega^{(1)}_{3}}$ & $\boldsymbol{\omega^{(1)}_{4}}$ & $\boldsymbol{\omega^{(1)}_{5}}$ & $\boldsymbol{\omega^{(1)}_{6}}$ & $\boldsymbol{\omega^{(1)}_{7}}$ & $\boldsymbol{\omega^{(1)}_{8}}$ \\
\midrule
${ g_{1t}(x=\text{\scriptsize Img0})}$& { 27.53} & { 8.98} & { -14.04} & { 6.89} & { 14.07} & { -22.24} & { 25.85} & {-27.13}\\
${ f_{1t}(x=\text{\scriptsize Img0})}$& { 27.53} & { 8.98} & { 0.0} & { 6.89} & { 14.07} & { 0.0} & { 25.85} & { 0.0}\\
${ \text{exp}[f_{1t}(x=\text{\scriptsize Img0})]}$ & \scriptsize{9.03e+11} & \scriptsize{7.94e+03} & \scriptsize{1.0} & \scriptsize{1.0} & \scriptsize{1.28e+06} & \scriptsize{1.0} & \scriptsize{1.68e+11} & {\scriptsize1.0}   \\
 ${\scriptstyle Q_{F_1|{X}}(\boldsymbol{\omega^{(1)}_{n}}|\text{Img0})}$ & \textbf{0.843} & { 0.0} & { 0.0} & { 0.0} & { 0.0} & {0.0} & \textbf{0.157} & { 0.0}   \\
\midrule
${ g_{1t}(x=\text{\scriptsize Img1})}$& -27.51 & -9.81 & 12.96 & -5.87 & -13.05 & 21.68 & -26.02 & 27.61\\
${ f_{1t}(x=\text{\scriptsize Img1})}$& 0.0 & 0.0 & 12.96 & 0.0 & 0.0 & 21.68 & 0.0 & 27.61\\
${ \text{exp}[f_{1t}(x=\text{\scriptsize Img1})]}$ & \scriptsize{1.0} & \scriptsize{1.0} & \scriptsize{4.25e+05} & \scriptsize{1.0} & \scriptsize{1.0} & \scriptsize{2.60e+09} & \scriptsize{1.0} & {\scriptsize 9.79e+11}   \\
 ${\scriptstyle Q_{F_1|{X}}(\boldsymbol{\omega^{(1)}_{t}}|\text{Img1})}$ & 0.0 & 0.0 & 0.0 & 0.0 & 0.0 & \textbf{0.003} & 0.0 & \textbf{0.997}   \\
\midrule
${ g_{1t}(x=\text{\scriptsize Img2})}$& -2.69 & 26.28 & 24.64 & -28.39 & -20.44 & 12.23 & 10.13 & -6.84\\
${ f_{1t}(x=\text{\scriptsize Img2})}$& 0.0 & 26.28 & 24.64 & 0.0 & 0.0 & 12.23 & 10.13 & 0.0\\
${ \text{exp}[f_{1t}(x=\text{\scriptsize Img2})]}$ & \scriptsize{1.0} & \scriptsize{2.58e+11} & \scriptsize{5.02e+10} & \scriptsize{1.0} & \scriptsize{1.0} & \scriptsize{2.04e+05} & \scriptsize{2.50e+04} & {\scriptsize 1.0}   \\
 ${\scriptstyle Q_{F_1|{X}}(\boldsymbol{\omega^{(1)}_{t}}|\text{Img2})}$ & 0.0 & \textbf{0.837} & \textbf{0.163} & 0.0 & 0.0 & 0.0 & 0.0 & 0.0   \\
\midrule
${ g_{1t}(x=\text{\scriptsize Img3})}$& 4.71 & -26.20 & -25.65 & 28.91 & 20.93 & -13.18 & -8.67 & 4.97\\
${ f_{1t}(x=\text{\scriptsize Img3})}$& 4.71 & 0.0 & 0.0 & 28.91 & 20.93 & 0.0 & 0.0 & 4.97\\
${ \text{exp}[f_{1t}(x=\text{\scriptsize Img3})]}$ & \scriptsize{1.11e+02} & \scriptsize{1.0} & \scriptsize{1.0} & \scriptsize{3.59e+12} & \scriptsize{1.22e+09} & \scriptsize{1.0} & \scriptsize{1.0} & {\scriptsize 1.44e+02}   \\
 ${\scriptstyle Q_{F_1|{X}}(\boldsymbol{\omega^{(1)}_{t}}|\text{Img3})}$ & 0.0 & 0.0 & 0.0 & \textbf{1.0} & 0.0 & 0.0 & 0.0 & 0.0   \\
\bottomrule
\end{tabular}
\begin{tablenotes}
            \item $g_{1t}({x})$ and $f_{1t}({x})$ denote the linear output and the activation, respectively, where $g_{1t}({x}) = \langle \boldsymbol{\omega}^{(1)}_{t}, {x} \rangle + b_{1t}$ and $f_{1t} = \sigma_1[g_{1t}({x})]$.
\end{tablenotes}
\end{threeparttable}
\end{small}
\end{center}
\vskip -0.1in
\end{table}

Second, we demonstrate the Gibbs probability measure $Q_{F_1}$ for ${f_1}$.
Based on Equation \ref{Gibbs_f1}, we summarize $Q_{F_1}(\boldsymbol{\omega}^{(1)}_{t})$ given the four synthetic images (Figure \ref{fig_iid_images1}) in Table \ref{fc_gibbs}, and find that $Q_{F_1}$ correctly measures the probability of the outcomes $\{\boldsymbol{\omega}^{(1)}_{t}\}_{t=1}^8$ occurring in the experiment based on the relation between $\boldsymbol{\omega}^{(1)}_{t}$ and ${x}$.
For example, $\boldsymbol{\omega}^{(1)}_{1}$ correctly describes the feature of Img0, thus it has the highest linear output and activation ($g_{11}({x}) = 27.53$ and $f_{11}({x}) = 27.53$), thereby $Q_{F|{X}}(\boldsymbol{\omega^{(1)}_{1}}|\text{Img0}) = 0.843$.
As a comparison, since $\boldsymbol{\omega}^{(1)}_{8}$ oppositely describes the feature of Img0, it has the lowest linear output and activation ($g_{18}({x}) = -27.13$ and $f_{18}({x}) = 0.0$), thereby $Q_{F|{X}}(\boldsymbol{\omega^{(1)}_{8}}|\text{Img0}) = 0.0$.
 
In summary, we demonstrate the probability space $(\Omega$, $\mathcal{T}, Q_F)$ for a fully connected layer based on the synthetic dataset. 
The learned weights of neurons generate a sample space $\Omega$ consisting of multiple possible outcomes/features occurring in the input.
The energy of each possible outcomes equals the negative activation of each neuron.
The Gibbs function $Q_F$ quantifies the probability of outcomes occurring in the input.

\subsubsection{The equivalence between SGD and the first order approximation}
\label{GD_Approx}

If an arbitrary function ${f}$ is differentiable at point ${p}^* \in {\mathbb{R}}^{N}$ and its differential is represented by the Jacobian matrix $\nabla_{p}{f}$, the first order approximation of ${f}$ near the point ${p}$ can be formulated as 
\begin{equation} 
{f}({p}) - {f}({p}^*) = (\nabla_{{p}^*}{f}) \cdot ({p} - {p}^*) + o(||{p} - {p}^*||),
\end{equation}
where $o(||{p} - {p}^*||)$ is a quantity that approaches zero much faster than $||{p} - {p}^*||$ approaches zero.

Based on the first order approximation \cite{GD_Taylor}, the activations in $f_2$ and $f_1$ in the $\text{MLP} = \{x, f_1, f_2, f_Y\}$ can be expressed as follows:
\begin{equation} 
\label{neuron_taylor0}
\begin{split}
{f_2}[{f_1}, {\theta}_{j+1}(2)] &\approx {f_2}[{f_1}, {\theta}_{j}(2)] + (\nabla_{{\theta}_{j}(2)} {f_2}) \cdot [{\theta}_{j+1}(2) - {\theta}_{j}(2)] \\
{f_1}[{x}, {\theta}_{j+1}(1)] &\approx {f_1}[{x}, {\theta}_{j}(1)] + (\nabla_{{\theta}_{j}(1)} {f_1}) \cdot [{\theta}_{j+1}(1) -{\theta}_{j}(1)], \\
\end{split}
\end{equation}
where ${f_2}[{f_1}, {\theta}_{j+1}(2)]$ are the activations of the second hidden layer based on the parameters of $f_2$ learned in the $j+1$th iteration, i.e., ${\theta}_{j+1}(2)$, given the activations of the first hidden layer, i.e., ${f_1}$.
In addition, the definitions of ${f_2}[{f_1}, {\theta}_{j}(2)]$, ${f_1}({x}, {\theta}_{j+1}(1))$, and ${f_1}({x}, {\theta}_{j}(1))$ are the same as ${f_2}[{f_1}, {\theta}_{j+1}(2)]$.

Since ${f_2} = \{f_{2k} = \sigma_2(\sum_{t=1}^T\omega^{(2)}_{tk} \cdot f_{1t} + b_{2k})\}_{k=1}^K$ has $K$ neurons and each neuron has $T+1$ parameters, namely ${\theta}(2) = \{\omega^{(2)}_{1k}; \cdots; \omega^{(2)}_{Tk}; b_{2k}\}_{k=1}^K$, the dimension of $\nabla_{{\theta}_j(2)} {f_2}$ is equal to $K \times (T+1)$ and $\nabla_{{\theta}_j(2)} {f_2}$ can be expressed as
\begin{equation} 
\begin{split}
\nabla_{{\theta}_j(2)} {f_2} =  (\nabla_{\sigma_2} {f_2}) \cdot [{f_1}; 1]^T
\end{split}
\end{equation}
where $\nabla_{\sigma_2} {f_2} = \frac{\partial {f_2}[{f_1}, {\theta}_t(2)]}{\partial \sigma_2}$. 
Substituting $(\nabla_{\sigma_2} {f_2}) \cdot [{f_1}; 1]^T$ for $\nabla_{{\theta}_j(2)} {f_2}$ in Equation \ref{neuron_taylor0}, we derive 
\begin{equation}
\label{neuron_taylor}
\begin{split}
{f_2}[{f_1}, {\theta}_{j+1}(2)] \approx &{f_2}[{f_1}, {\theta}_j(2)] + (\nabla_{\sigma_2} {f_2}) \cdot [{f_1}; 1]^T \cdot {\theta}_{j+1}(2) - (\nabla_{\sigma_2} {f_2}) \cdot [{f_1}; 1]^T \cdot {\theta}_j(2)
\end{split}
\end{equation}

If we only consider a single neuron, e.g., $f_{2k}$, we define ${\theta}_{j+1}(2k) = [\omega^{(1)}_{1k}; \cdots; \omega^{(1)}_{Tk}; b_{2k}]$ and ${\theta}_j(2k) = [\omega'^{(1)}_{1k}; \cdots; \omega'^{(1)}_{Tk}; b'_{2k}]$, thus $[{f_1}; 1]^T \cdot {\theta}_{j+1}(2k) = \sum_{t=1}^T\omega^{(2)}_{tk} \cdot f_{1t} + b_{2k}$.
As a result, for a single neuron, Equation \ref{neuron_taylor} can be expressed as 
\begin{equation} 
\label{neuron_taylor2}
\begin{split}
{f_{2k}}[{f_1}, {\theta}_{j+1}(2k)] \approx \underbrace{(\nabla_{\sigma_2} f_{2k}) \cdot [\sum_{t=1}^T\omega^{(2)}_{tk} \cdot f_{1t} + b_{2k}]}_{\text{Approximation}} + \underbrace{f_{2k}[{f_1}, {\theta}_j(2k)] - (\nabla_{\sigma_2} f_{2k}) \cdot [\sum_{t=1}^T\omega'^{(2)}_{tk} \cdot f_{1t} + b'_{2k}]}_{\text{Bias}}
\end{split}
\end{equation}

Equation \ref{neuron_taylor2} indicates that ${f_{2k}}[{f_1}, {\theta}_{j+1}(2k)]$ can be reformulated as two components:  approximation and bias.
Since $\nabla_{\sigma_2} f_{2k} = \frac{\partial f_{2k}[{f_1}, {\theta}_j(2)]}{\partial \sigma_2}$ is only related to ${f_1}$ and ${\theta}_j(2)$, it can be regarded as a constant with respect to ${\theta}_{j+1}(2)$.
The bias component also does not contain any parameters in the $(j+1)$th training iteration. 

In summary, ${f_{2k}}({f_1}, {\theta}_{j+1}(2k))$ can be reformulated as
\begin{equation} 
\label{neuron_taylor3}
\begin{split}
{f_{2k}}({f_1}, {\theta}_{j+1}(2k)) & \approx C_1 \cdot [\sum_{t=1}^T\omega^{(2)}_{tk} \cdot f_{1t} + b_{2k}]  + C_2 \\
\end{split}
\end{equation}
where $C_1 = \nabla_{\sigma_2} f_{2k}$ and $C2 = f_{2k}({f_1}, {\theta}_j(2k)) - (\nabla_{\sigma_2} f_{2k}) \cdot [\sum_{t=1}^T\omega^{(2)}_{tk} \cdot f_{1t} + b^*_{2k}]$.
Similarly, the activations in the first hidden layer ${f_1}$ also can be formulated as the approximation.
To validate the approximation, we only need prove ${\theta}_{j+1}(2) - {\theta}_{j}(2)$ approaching zero, which can be guaranteed by SGD.

\pagebreak
Given the $\text{MLP} = \{{x; f_1; f_2; f_Y}\}$ and the empirical risk $\hat{\mathcal{L}}^{\ell}_{\mathcal{S}}(h)$, 
SGD aims to optimize the parameters of the MLP through minimizing $\hat{\mathcal{L}}^{\ell}_{\mathcal{S}}(h)$ \cite{backpropagation}.
\begin{equation} 
\label{gd_theta}
{\theta}_{t+1} = {\theta}_t - \alpha \nabla_{\theta_t} \hat{\mathcal{L}}^{\ell}_{\mathcal{S}}(h),
\end{equation}
where $\nabla_{\theta_t} \hat{\mathcal{L}}^{\ell}_{\mathcal{S}}(h)$ denotes the Jacobian matrix of $\hat{\mathcal{L}}^{\ell}_{\mathcal{S}}(h)$ with respect to ${\theta}_t$ at the $t$th iteration, and $\alpha > 0$ denotes the learning rate.
Since the functions of all the layers are differentiable, the Jacobian matrix of $\hat{\mathcal{L}}^{\ell}_{\mathcal{S}}(h)$ with respect to the parameters of the $i$th hidden layer, i.e., $\nabla_{\theta(i)} \hat{\mathcal{L}}^{\ell}_{\mathcal{S}}(h)$, can be expressed as
\begin{equation} 
\label{jacobian_theta}
\begin{split}
\nabla_{{\theta}(y)} \hat{\mathcal{L}}^{\ell}_{\mathcal{S}}(h) &= \nabla_{{f_Y}} \hat{\mathcal{L}}^{\ell}_{\mathcal{S}}(h)\nabla_{{\theta_Y}}f_Y\\
\nabla_{{\theta}(2)} \hat{\mathcal{L}}^{\ell}_{\mathcal{S}}(h) &= \nabla_{{f_Y}} \hat{\mathcal{L}}^{\ell}_{\mathcal{S}}(h)\nabla_{f_2}f_Y \nabla_{\theta_2} f_2\\
\nabla_{{\theta}(1)} \hat{\mathcal{L}}^{\ell}_{\mathcal{S}}(h) &= \nabla_{{f_Y}} \hat{\mathcal{L}}^{\ell}_{\mathcal{S}}(h) \nabla_{{f_2}} f_Y \nabla_{{f_1}} f_2 \nabla_{\theta_1} f_1,
\end{split}
\end{equation}
where ${\theta}(i)$ denote the parameters of the $i$th layer.
Equation \ref{gd_theta} and \ref{jacobian_theta} indicate that ${\theta}(i)$ can be learned as
\begin{equation} 
{\theta}_{t+1}(i) = {\theta}_t(i) - \alpha [\nabla_{\theta_t(i)} \hat{\mathcal{L}}^{\ell}_{\mathcal{S}}(h)].
\end{equation}
Table \ref{backpropagation_dnn} summarizes SGD training procedure for the MLP shown in Figure \ref{fig_mlp_app}.

SGD minimizing $\hat{\mathcal{L}}^{\ell}_{\mathcal{S}}(h)$ indicates $\nabla_{\theta_t(i)} \hat{\mathcal{L}}^{\ell}_{\mathcal{S}}(h)$ converging to zero, thereby ${\theta}_{t+1}(i) - {\theta}_t(i)$ converging to zero.

\begin{table}
\caption{One iteration of SGD training procedure for the MLP}
\label{backpropagation_dnn}
\vskip 0.15in
\begin{center}
\begin{small}
\begin{threeparttable}
\begin{tabular}{ccccccc}
\toprule
Layer & Gradients $\nabla_{\theta(i)} \hat{\mathcal{L}}^{\ell}_{\mathcal{S}}(h)$ & & Parameters & &Activations & \\
\midrule
${f_Y}$		&  ${\scriptstyle \nabla_{\boldsymbol{\theta}(y)} \hat{\mathcal{L}}^{\ell}_{\mathcal{S}}(h)}$ & $\downarrow$& ${\scriptstyle {\theta}_{t+1}(Y) = {\theta}_{t+1}(Y) - \alpha[\nabla_{{\theta}(y)} \hat{\mathcal{L}}^{\ell}_{\mathcal{S}}(h)]}$ & $\uparrow$ & ${\scriptstyle {f_Y}({f_2}, {\theta}_{t+1}(Y))}$ & $\uparrow$ \\
${f_2}$		& ${\scriptstyle \nabla_{\boldsymbol{\theta}(2)} \hat{\mathcal{L}}^{\ell}_{\mathcal{S}}(h)} $ & $\downarrow$& ${\scriptstyle {\theta}_{t+1}(2) = {\theta}_{t+1}(2) - \alpha[\nabla_{{\theta}(2)} \hat{\mathcal{L}}^{\ell}_{\mathcal{S}}(h)]}$ & $\uparrow$ & ${\scriptstyle {f_2}({f_1}, {\theta}_{t+1}(2))}$ & $\uparrow$ \\
${f_1}$    	& ${\scriptstyle \nabla_{\boldsymbol{\theta}(1)} \hat{\mathcal{L}}^{\ell}_{\mathcal{S}}(h)} $ & $\downarrow$& ${\scriptstyle {\theta}_{t+1}(1) = {\theta}_{t+1}(1) - \alpha[\nabla_{{\theta}(1)} \hat{\mathcal{L}}^{\ell}_{\mathcal{S}}(h)]}$ & $\uparrow$ & ${\scriptstyle {f_1}({x}, {\theta}_{t+1}(1))}$ & $\uparrow$\\
${x}$    	        & ---	& & --- & & ---\\

\bottomrule
\end{tabular}
\begin{tablenotes}
            \item The up-arrow and down-arrow indicate the order of gradients and parameters(activations) update, respectively. 
\end{tablenotes}
\end{threeparttable}
\end{small}
\end{center}
\vskip -0.1in
\end{table}

\subsection{The derivation of the marginal distribution $Q(F_Y|{X})$}
\label{posterior_MLP}

Since the entire architecture of the $\text{MLP} = \{{x; f_1; f_2; f_Y}\}$ in Figure \ref{fig_mlp_app} corresponds to a joint distribution $Q(F_Y; F_2; F_1|{X}) = Q(F_Y|F_2)P(F_2|F_1)P(F_1|{X})$, the marginal distribution $Q(F_Y|{X})$ can be formulated as 
\begin{equation} 
\begin{split}
Q_{F_Y|X}(l|{x}_i) &= \sum_{k=1}^K \sum_{t=1}^T Q(F_Y = l, F_2 = k, F_1=t|X={x}_i) \\
&= \sum_{k=1}^K Q_{F_Y|F_2}(l|k) \sum_{t=1}^T Q_{F_2|F_1}(k|t)Q_{F_1|X}(t|{x}_{i}). \\
\end{split}
\end{equation}

Based on the definition of the Gibbs probability measure (Equation \ref{Gibbs_f}), we have
\begin{equation} 
Q_{F_1|X}(t|{x}_{i}) = \frac{1}{Z_{{F_1}}}\text{exp}(f_{1t}) = \frac{1}{Z_{{F_1}}}\text{exp}[\sigma_1(\langle {\boldsymbol{\omega}'}^{(1)}_{t}, {x}'_i \rangle)],
\end{equation}
where ${\boldsymbol{\omega}'}^{(1)}_{t} = [{\boldsymbol{\omega}}^{(1)}_{t}, b_{1n}]$ and ${x}'_i = [{x}_i, 1]$, i.e., $\langle {\boldsymbol{\omega}'}^{(1)}_{t}, {x}'_i \rangle = \langle {\boldsymbol{\omega}}^{(1)}_{t}, {x}_i \rangle + b_{1n}$.

Similarly, we have
\begin{equation} 
Q_{F_2|F_1}(k|t) = \frac{1}{Z_{{F_2}}}\text{exp}(f_{2k}) = \frac{1}{Z_{{F_2}}}\text{exp}[\sigma_2(\langle {\boldsymbol{\omega}'}^{(2)}_{k}, {f}'_1 \rangle)],
\end{equation}
where ${f}_1 = \{f_{1t}\}_{t=1}^T = \{\sigma_1(\langle {\boldsymbol{\omega}'}^{(1)}_{t}, {x}'_i \rangle)\}_{t=1}^T$, ${\boldsymbol{\omega}'}^{(2)}_{k} = [{\boldsymbol{\omega}}^{(2)}_{k}, b_{2k}]$ and ${f}'_1 = [{f}_1, 1]$, i.e., $\langle {\boldsymbol{\omega}'}^{(2)}_{k}, {f}'_1 \rangle = \langle {\boldsymbol{\omega}}^{(2)}_{k}, {f}_1 \rangle + b_{2k}$.
Therefore, we have 
\begin{equation} 
\begin{split}
\sum_{t=1}^T Q_{F_2|F_1}(k|t)Q_{F_1|X}(t|{x}_{i}) = \frac{1}{Z_{{F_2}}}\frac{1}{Z_{{F_1}}}\sum_{t=1}^T\text{exp}[\sigma_2(\langle {\boldsymbol{\omega}'}^{(2)}_{k}, {f}'_1 \rangle)]\text{exp}[\sigma_1(\langle {\boldsymbol{\omega}'}^{(1)}_{t}, {x}'_i \rangle)]. \\
\end{split}
\end{equation}

Since $\langle {\boldsymbol{\omega}'}^{(2)}_{k}, {f}'_1 \rangle = \langle {\boldsymbol{\omega}}^{(2)}_{k}, {f}_1 \rangle + b_{2k} = \sum_{t=1}^T\omega^{(2)}_{kt}f_{1t} + b_{2k}$ is a constant with respect to $t$, we have
\begin{equation} 
\sum_{t=1}^T Q_{F_2|F_1}(k|t)Q_{F_1|X}(t|{x}_{i}) = \frac{1}{Z_{{F_2}}}\frac{1}{Z_{{F_1}}}\text{exp}[\sigma_2(\langle {\boldsymbol{\omega}'}^{(2)}_{k}, {f}'_1 \rangle)]\sum_{t=1}^T\text{exp}[\sigma_1(\langle {\boldsymbol{\omega}'}^{(1)}_{t}, {x}'_i \rangle)]. 
\end{equation}
In addition, $\sum_{t=1}^T\text{exp}[\sigma_1(\langle {\boldsymbol{\omega}'}^{(1)}_{t}, {x}'_i \rangle)] = Z_{F_1}$, thus we have 
\begin{equation}
\sum_{t=1}^T Q_{F_2|F_1}(k|t)Q_{F_1|X}(t|{x}_{i}) = \frac{1}{Z_{{F_2}}}\text{exp}[\sigma_2(\langle {\boldsymbol{\omega}'}^{(2)}_{k}, {f}'_1 \rangle)].
\end{equation}
Therefore, we can simplify $Q_{F_Y|X}(l|{x}_i)$ as
\begin{equation}
\begin{split}
Q_{F_Y|X}(l|{x}_i) &= \sum_{k=1}^K Q_{F_Y|F_2}(l|k) \sum_{t=1}^T Q_{F_2|F_1}(k|t)Q_{F_1|X}(t|{x}_{i}) \\
&= \sum_{k=1}^K Q_{F_Y|F_2}(l|k) \frac{1}{Z_{{F_2}}}\text{exp}[\sigma_2(\langle {\boldsymbol{\omega}'}^{(2)}_{k}, {f}'_1 \rangle)].\\
\end{split}
\end{equation}
Similarly, since $Q_{F_Y|F_2}(l|k) = \frac{1}{Z_{{F_Y}}}\text{exp}[\sigma_3(\langle {\boldsymbol{\omega}}^{(3)}_{l}, {f}_2 \rangle + b_{yl})]$ and $\langle {\boldsymbol{\omega}}^{(3)}_{l}, {f}_2 \rangle = \sum_{k=1}^K\omega^{(3)}_{lk}f_{2k}$ is also a constant with respect to $k$, we can derive 
\begin{equation}
\begin{split}
Q_{F_Y|X}(l|{x}_i) = Q_{F_Y|F_2}(l|k) = \frac{1}{Z_{F_Y}}\text{exp}[\sigma_3(\langle {\boldsymbol{\omega}}^{(3)}_{l}, {f}_2 \rangle + b_{yl})].
\end{split}
\end{equation}

In addition, since ${f}_2 = \{f_{2k}\}_{k=1}^K = \{\sigma_2(\langle {\boldsymbol{\omega}}^{(2)}_{k}, {f}_1 \rangle + b_{2k}) \}_{k=1}^K$, we can extend $Q_{F_Y|X}(l|{x}_i)$ as 
\begin{equation}
\begin{split}
Q_{F_Y|X}(l|{x}_i) &= Q_{F_Y|F_2}(l|k) = \frac{1}{Z_{F_Y}}\text{exp}[\sigma_3(\langle {\boldsymbol{\omega}}^{(3)}_{l}, {f}_2 \rangle + b_{yl})]\\
&= \frac{1}{Z_{F_Y}}\text{exp}[\sigma_3(\langle {\boldsymbol{\omega}}^{(3)}_{l}, \left( \begin{array}{c} \sigma_2(\langle {\boldsymbol{\omega}}^{(2)}_{1}, {f}_1 \rangle + b_{21}) \\ \vdots \\ \sigma_2(\langle {\boldsymbol{\omega}}^{(2)}_{K}, {f}_1 \rangle + b_{2K}) \end{array} \right) \rangle + b_{yl})].
\end{split}
\end{equation}

Since ${f}_1 = \{f_{1t}\}_{t=1}^T = \{\sigma_1(\langle {\boldsymbol{\omega}}^{(1)}_{t}, {x}_i \rangle + b_{1n})\}_{t=1}^T$, we can further extend $Q_{F_Y|X}(l|{x}_i)$ as 
\begin{equation}
\begin{split}
Q_{F_Y|X}(l|{x}_i)
&= \frac{1}{Z_{F_Y}}\text{exp}[\sigma_3(\langle {\boldsymbol{\omega}}^{(3)}_{l}, \left( \begin{array}{c} \sigma_2(\langle {\boldsymbol{\omega}}^{(2)}_{1}, \left( \begin{array}{c} \sigma_1(\langle {\boldsymbol{\omega}}^{(1)}_{1}, {x}_i \rangle + b_{11}) \\ \vdots \\ \sigma_1(\langle {\boldsymbol{\omega}}^{(1)}_{1}, {x}_i \rangle + b_{11}) \end{array} \right) \rangle + b_{21}) \\ \vdots \\ \sigma_2(\langle {\boldsymbol{\omega}}^{(2)}_{K}, \left( \begin{array}{c} \sigma_1(\langle {\boldsymbol{\omega}}^{(1)}_{T}, {x}_i \rangle + b_{1T}) \\ \vdots \\ \sigma_1(\langle {\boldsymbol{\omega}}^{(1)}_{T}, {x}_i \rangle + b_{1T}) \end{array} \right) \rangle + b_{2K}) \end{array} \right) \rangle + b_{yl})].\\
&= \frac{1}{Z_{\text{MLP}}(x_i)}\text{exp}[{f_{yl}(f_2(f_1({x}_i)))}]
\end{split}
\end{equation}

Overall, we prove that $Q_{F_Y|X}(l|{x}_i)$ is also a Gibbs distribution and it can be expressed as 
\begin{equation}
\label{posterior_dnn} 
Q_{F_Y|X}(l|{x}_i) = \frac{1}{Z_{\text{MLP}}({x}_i)}\text{exp}[f_{yl}(f_2(f_1({x}_i)))].
\end{equation}
where $E_{yl}(x_i) = - f_{yl}(f_2(f_1({x}_i)))$ is the energy function of $l \in \mathcal{Y}$ given $x_i$ and the partition function $Z_{\text{MLP}}({x}_i) = \sum_{l=1}^L\sum_{k=1}^K\sum_{t=1}^TQ(F_Y,F_2,F_1|X=x_i) = \sum_{l=1}^L\text{exp}[f_{yl}(f_2(f_1({x}_i)))]$.

\clearpage

\subsection{$\text{KL}[P^*(F_Y|X, Y)||Q(F_Y|X)]$ approachs $\text{KL}[Q(F_Y|X)||P^*(F_Y|X, Y)]$}
\label{KL_asymmetric}

This appendix demonstrates $\text{KL}[P^*(F_Y|X, Y)||Q(F_Y|X)] \rightarrow \text{KL}[Q(F_Y|X)||P^*(F_Y|X, Y)]$ because $d \rightarrow 0$ during minimizing $\hat{\mathcal{L}}^{\ell}_{\mathcal{S}}(h)$.

Recall $d = \text{KL}[P^*(F_Y|X, Y)||Q(F_Y|X)] - \text{KL}[Q(F_Y|X)||P^*(F_Y|X, Y)]$,  the property of cross entropy, i.e., $\text{KL}[P||Q] = H(P,Q) - H(P)$, allows us to derive
\begin{equation} 
d = \underbrace{H[P^*(F_Y|X, Y), Q(F_Y|X)] - H[P^*(F_Y|X, Y)]}_{\text{KL}[P^*(F_Y|X, Y)||Q(F_Y|X)]} - \underbrace{\{H[Q(F_Y|X),P^*(F_Y|X, Y)] - H[Q(F_Y|X)]\}}_{\text{KL}[Q(F_Y|X)||P^*(F_Y|X, Y)]}.
\end{equation}
where $H(P)$ denotes the entropy of $P$.
Since $P^*(F_Y|X, Y)$ is one-hot, namely that if $l = y_i$, then $P^*_{F_Y|X, Y}(l|x_i, y_i)  = 1$, otherwise $P^*_{F_Y|X, Y}(l|x_i, y_i) = 0$,  we have 
\begin{equation} 
\begin{split}
H[P^*(F_Y|X, Y)] &= 0,\\
H[P^*(F_Y|X, Y), Q(F_Y|X)] &= -\text{log}Q_{F_Y|X}(y_i|x_i).
\end{split}
\end{equation}
Therefore, we can simplify $d$ as 
\begin{equation} 
d = -\text{log}Q_{F_Y|X}(y_i|x_i) + \sum_{l=1}^LQ_{F_Y|X}(l|x_i)\text{log}P^*_{F_Y|X, Y}(l|x_i,y_i) + H[Q(F_Y|X)].
\end{equation}
$P^*_{F_Y|X,Y}(l|x_i, y_i)=0$ induces $\text{log}P^*_{F_Y|X,Y}(l|x_i, y_i) = -\infty$. 
To guarantee $d$ is a valid value, we relax $P^*(F_Y|X, Y)$ as  
\begin{equation}
\hat{P}^*_{F_Y|X,Y}(l|x_i, y_i) =     \left\{ \begin{array}{rcl}
         1-(L-1) \varepsilon & \text{if} & l = y_i \\ 
         \varepsilon \;\; \;\; \;     & \text{if}  & l \neq y_i
                \end{array}\right.
\end{equation} 
where $\varepsilon$ is a small positive value. As a result, we have
\begin{equation}
\begin{split}
d = &-\text{log}Q_{F_Y|X}(y_i|x_i) + [1-Q_{F_Y|X}(y_i|x_i)](L-1)\text{log}\varepsilon + H[Q(F_Y|X=x_i)] \\
& + Q_{F_Y|X}(y_i|x_i)\text{log}[1-(L-1)\varepsilon].
\end{split}
\end{equation} 

Corollary 1 indicates that minimizing $\hat{\mathcal{L}}^{\ell}_{\mathcal{S}}(h)$ is equivalent to maximizing  $\sum_{i=1}^n\text{log}Q_{F_Y|X}(y_i|x_i)$. 
Since $Q_{F_Y|X}(y_i|x_i) \in [0, 1]$, maximizing  $\sum_{i=1}^n\text{log}Q_{F_Y|X}(y_i|x_i)$ makes $Q_{F_Y|X}(y_i|x_i)$ to approach 1.
As a result, $\text{log}Q_{F_Y|X}(y_i|x_i)$, $[1-Q_{F_Y|X}(y_i|x_i)](L-1)\text{log}\varepsilon$, and $H[Q(F_Y|X=x_i)]$ approach 0.
In addition, we have $\underset{\varepsilon \rightarrow 0}{\text{lim}}Q_{F_Y|X}(y_i|x_i)\text{log}[1-(L-1)\varepsilon]  = 0$.
In summary, $d$ converges to 0 when minimizing $\hat{\mathcal{L}}^{\ell}_{\mathcal{S}}(h)$.

\begin{figure}[!b]
\centering
\includegraphics[scale=0.55]{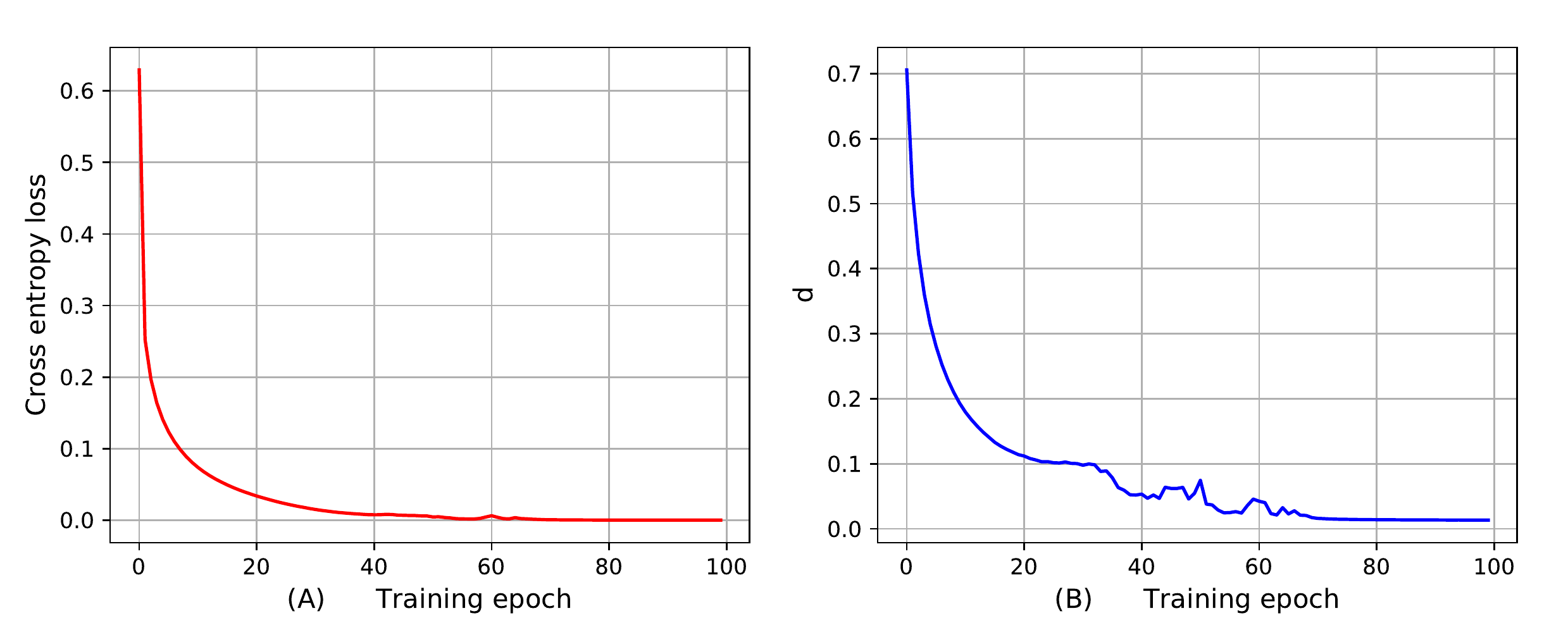}
\caption{\small{
(A) and (B) show the variations of the cross entropy loss and $d$ during 100 training epochs, respectively.
}}
\label{Img_d_0}
\end{figure}

Moreover, we train the MLP1 designed in Section \ref{simulations} on the MNIST dataset and visualize the variation of the cross entropy loss and $d$ during 100 epochs in Figure \ref{Img_d_0}. 
We can observe that $d$ shows a similar decreasing trend as the cross entropy loss and $d$ converges to zero when the cross entropy loss approaches zero. 
Since $d$ converges to zero, $\text{KL}[P^*(F_Y|X, Y)||Q(F_Y|X)] \rightarrow \text{KL}[Q(F_Y|X)||P^*(F_Y|X, Y)]$ during minimizing $\hat{\mathcal{L}}^{\ell}_{\mathcal{S}}(h)$.

\subsection{The derivation of the ELBO from Bayesian variational inference}
\label{elbo_app}

Given the samples $\mathcal{S}$, the $\text{MLP} = \{{x; f_1;f_2; f_Y}\}$, and the cross entropy loss $\hat{\mathcal{L}}^{\ell}_{\mathcal{S}}(h)$, 
recall the corresponding Bayesian variational inference (Equation \ref{mlp_vi}) as
\begin{equation}
\text{KL}[{\scriptstyle Q_{\mathcal{S}_x}(F_Y|X)}||{\scriptstyle P^*_{\mathcal{S}}(F_Y|X, Y)}] = \underset{Q(F_Y|X)}{\boldsymbol{E}}\text{log}{\scriptstyle Q_{\mathcal{S}_x}(F_Y|X)} -  \underset{Q(F_Y|X)}{\boldsymbol{E}} \text{log}{\scriptstyle P^*_{\mathcal{S}}(F_Y|X, Y)}.
\end{equation}
Since $P^*_{\mathcal{S}}(F_Y|X,Y) = P^*_{\mathcal{S}}(F_Y,Y|X)/P_{\mathcal{S}}(Y|X)$, we have
\begin{equation}
\text{KL}[{\scriptstyle Q_{\mathcal{S}_x}(F_Y|X)}||{\scriptstyle P^*_{\mathcal{S}}(F_Y|X, Y)}] 
= \underset{Q(F_Y|X)}{\boldsymbol{E}} \text{log}{\scriptstyle Q_{\mathcal{S}_x}(F_Y|X)} - \underset{Q(F_Y|X)}{\boldsymbol{E}} [\text{log}{\scriptstyle P^*_{\mathcal{S}}(F_Y,Y|X)} - \text{log}{\scriptstyle P_{\mathcal{S}}(Y|X)}].
\end{equation}
Since $\text{log}P_{\mathcal{S}}(Y|X)$ is constant with respect to $Q(F_Y|X)$, we have
\begin{equation}
\text{KL}[{\scriptstyle Q_{\mathcal{S}_x}(F_Y|X)}||{\scriptstyle P^*_{\mathcal{S}}(F_Y|X, Y)}] 
= \underset{Q(F_Y|X)}{\boldsymbol{E}} \text{log}{\scriptstyle Q_{\mathcal{S}_x}(F_Y|X)} - \underset{Q(F_Y|X)}{\boldsymbol{E}} \text{log}{\scriptstyle P^*_{\mathcal{S}}(F_Y,Y|X)} + \text{log}{\scriptstyle P_{\mathcal{S}}(Y|X)}.
\end{equation}
Since $P^*_{\mathcal{S}}(F_Y,Y|X) = P_{\mathcal{S}}(Y|F_Y, X)P^*_{\mathcal{S}_x}(F_Y|X)$, we have
\begin{equation}
\begin{split}
\text{KL}[{\scriptstyle Q_{\mathcal{S}_x}(F_Y|X)}||{\scriptstyle P^*_{\mathcal{S}}(F_Y|X, Y)}] 
= &\underset{Q(F_Y|X)}{\boldsymbol{E}} \text{log}{\scriptstyle Q_{\mathcal{S}_x}(F_Y|X)} - \underset{Q(F_Y|X)}{\boldsymbol{E}} \text{log}{\scriptstyle P^*_{\mathcal{S}_x}(F_Y|X)} - \underset{Q(F_Y|X)}{\boldsymbol{E}} \text{log}{\scriptstyle P_{\mathcal{S}}(Y|F_Y, X)}\\
&+ \text{log}{\scriptstyle P_{\mathcal{S}}(Y|X)}.
\end{split}
\end{equation}
Since $\underset{Q(F_Y|X)}{\boldsymbol{E}} \text{log}{\scriptstyle Q_{\mathcal{S}_x}(F_Y|X)} - \underset{Q(F_Y|X)}{\boldsymbol{E}} \text{log}{\scriptstyle P^*_{\mathcal{S}_x}(F_Y|X)} = \text{KL}[{\scriptstyle Q_{\mathcal{S}_x}(F_Y|X)}|| {\scriptstyle P^*_{\mathcal{S}_x}(F_Y|X)}]$, we have
\begin{equation}
\begin{split}
\text{KL}[{\scriptstyle Q_{\mathcal{S}_x}(F_Y|X)}||{\scriptstyle P^*_{\mathcal{S}}(F_Y|X, Y)}] 
= \text{KL}[{\scriptstyle Q_{\mathcal{S}_x}(F_Y|X)}|| {\scriptstyle P^*_{\mathcal{S}_x}(F_Y|X)}] - \underset{Q(F_Y|X)}{\boldsymbol{E}} \text{log}{\scriptstyle P_{\mathcal{S}}(Y|F_Y, X)} + \text{log}{\scriptstyle P_{\mathcal{S}}(Y|X)}.
\end{split}
\end{equation}
Finally, we can derive the ELBO as
\begin{equation}
\text{log}P_{\mathcal{S}}(Y|X) = \underbrace{\underset{Q(F_Y|X)}{\boldsymbol{E}}\text{log} {\scriptstyle P_{\mathcal{S}}(Y|F_Y,X)}-\text{KL}[{\scriptstyle Q_{\mathcal{S}_x}(F_Y|X)}|| {\scriptstyle P^*_{\mathcal{S}_x}(F_Y|X)}]}_{\text{ELBO}} + \text{KL}[{\scriptstyle Q_{\mathcal{S}_x}(F_Y|X)}||{\scriptstyle P^*_{\mathcal{S}}(F_Y|X, Y)}],
\end{equation}
where $P_{\mathcal{S}}(Y|X) = \prod_{i=1}^nP(Y=y_i|X=x_i)$, $P_{\mathcal{S}}(Y|F_Y,X) = \prod_{i=1}^nP(Y=y_i|F_Y, X=x_i)$, $Q_{\mathcal{S}_x}(F_Y|X) = \prod_{i=1}^nQ(F_Y|X=x_i)$, and $P^*_{\mathcal{S}_x}(F_Y|X) = \prod_{i=1}^nP^*(F_Y|X=x_i)$.

Again, since $\text{log}P_{\mathcal{S}}(Y|X)$ is constant with respect to $Q(F_Y|X)$, minimizing $\text{KL}[{\scriptstyle Q_{\mathcal{S}_x}(F_Y|X)}||{\scriptstyle P^*_{\mathcal{S}}(F_Y|X, Y)}]$ is equivalent to maximizing the ELBO.

\subsection{The variations of $\frac{1}{n}\sum_{i=1}^n\text{log}Z_{\text{MLP}}(x_i)$ and $\frac{1}{n}\sum_{i=1}^nE_{yy_i}(x_i)$ during minimizing $\hat{\mathcal{L}}^{\ell}_{\mathcal{S}}(h)$}
\label{energy_log_partition}

In this appendix, we show the variation of $\frac{1}{n}\sum_{i=1}^n\text{log}Z_{\text{MLP}}(x_i)$ and $\frac{1}{n}\sum_{i=1}^nE_{yy_i}(x_i)$ during minimizing $\hat{\mathcal{L}}^{\ell}_{\mathcal{S}}(h)$, and prove that $\frac{1}{n}\sum_{i=1}^n\text{log}Z_{\text{MLP}}(x_i)$ is not helpful to decrease the generalization bound because
$Z_{\text{MLP}}(x_i)$ is a constant with respect to $Q(F_Y|X)$.
 
\begin{figure}[!t]
\centering
\includegraphics[scale=0.55]{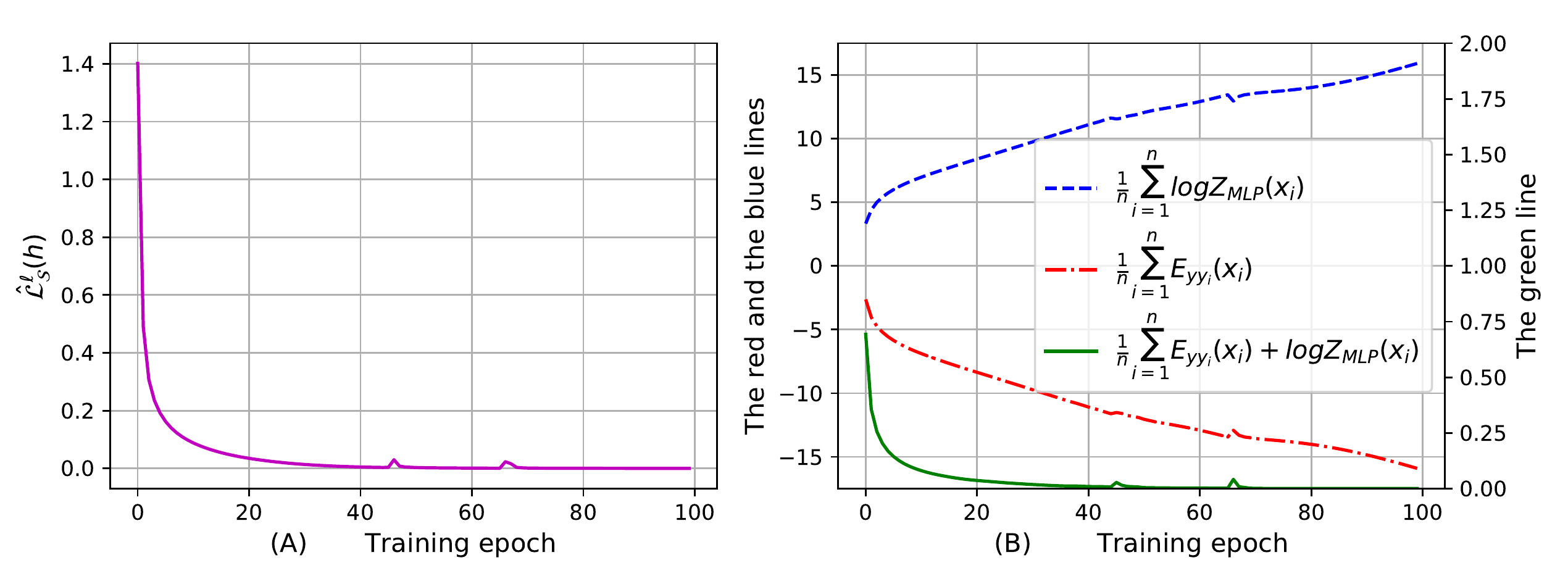}
\caption{
(A) shows the variation of $\hat{\mathcal{L}}^{\ell}_{\mathcal{S}}(h)$ during 100 training epochs.
(B) shows the variations of $\frac{1}{n}\sum_{i=1}^n\text{log}Z_{\text{MLP}}(x_i)$, $\frac{1}{n}\sum_{i=1}^nE_{yy_i}(x_i)$, and their summation during 100 training epochs.
}
\label{Img_bound_log_energy}
\end{figure}

We train MLP1 designed in Section \ref{simulations} on the MNIST dataset and visualize the variations of $\frac{1}{n}\sum_{i=1}^n\text{log}Z_{\text{MLP}}(x_i)$, $\frac{1}{n}\sum_{i=1}^nE_{yy_i}(x_i)$, and their summation during minimizing $\hat{\mathcal{L}}^{\ell}_{\mathcal{S}}(h)$ in Figure \ref{Img_bound_log_energy}.
We can observe that $\frac{1}{n}\sum_{i=1}^n\text{log}Z_{\text{MLP}}(x_i)$ is increasing and $\frac{1}{n}\sum_{i=1}^nE_{yy_i}(x_i)$ is decreasing.
Notably, their summation $\frac{1}{n}\sum_{i=1}^nE_{yy_i}(x_i) + \text{log}Z_{\text{MLP}}(x_i)$ has the same decreasing trend as $\frac{1}{n}\sum_{i=1}^nE_{yy_i}(x_i)$ and quickly converges to zero, which indicates
\begin{equation}
Q_{F_Y|X}(y_i|x_i) = \frac{1}{Z_{\text{MLP}}(x_i)}\text{exp}[-E_{yy_i}(x_i)] \rightarrow 1.
\end{equation}
Recall that $Z_{\text{MLP}}(x_i) = \sum_{l=1}^L \sum_{k=1}^K \sum_{t=1}^T Q(F_Y, F_2, F_1|X=x_i)$ sums up all the cases of $F_Y$. As a result, $Z_{\text{MLP}}(x_i)$ is a constant with respect to $Q(F_Y|X)$.
In other words, the functionality of $Z_{\text{MLP}}(x_i)$ is only to guarantee the validity of $Q(F_Y|X)$, thus it can be viewed as a constant for deriving the generalization bound.

As a result, the PAC-Bayesian generalization bound (Equation \ref{bound}) can be relaxed as 
\begin{equation}
\text{the PAC-Bayesian generalization bound} \propto \frac{1}{n}\sum_{i=1}^nE_{yy_i}(x_i).
\end{equation} 

\pagebreak
\subsection{Experiments on MLPs with the Fashion-MNIST dataset}
\label{mlp_fashin_mnist}

This appendix validates the PAC-Bayesian generalization bound for MLPs on the Fashion-MNIST dataset \cite{xiao2017fashion}.

\textbf{The bound can reflect the generalization of MLPs with different number of neurons.}

\begin{figure}[!b]
\centering
\includegraphics[scale=0.55]{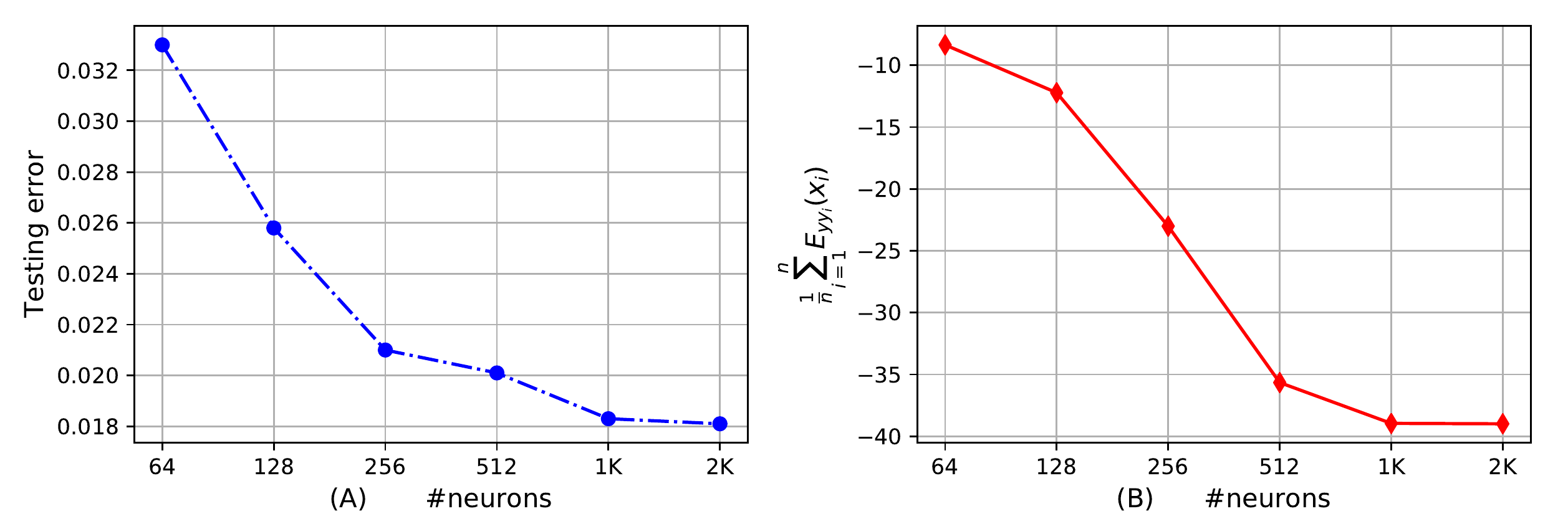}
\caption{
(A) and (B) show the testing error and the bound with different number of neurons, respectively.
}
\label{Img_generalization_architecture_A}
\end{figure}

We train the $\text{MLP} = \{{x; f_1;f_2; f_Y}\}$ with different number of neurons on the Fashion-MNIST dataset, and visualize the testing error and the bound in Figure \ref{Img_generalization_architecture_A}.
We can observe that the MLP achieves lower testing error as the number of neurons increases, and the variation of the bound is consistent with that of testing error.

\textbf{The bound can reflect the effect of SGD optimization on the generalization of MLPs.}

\begin{figure}[!t]
\centering
\includegraphics[scale=0.55]{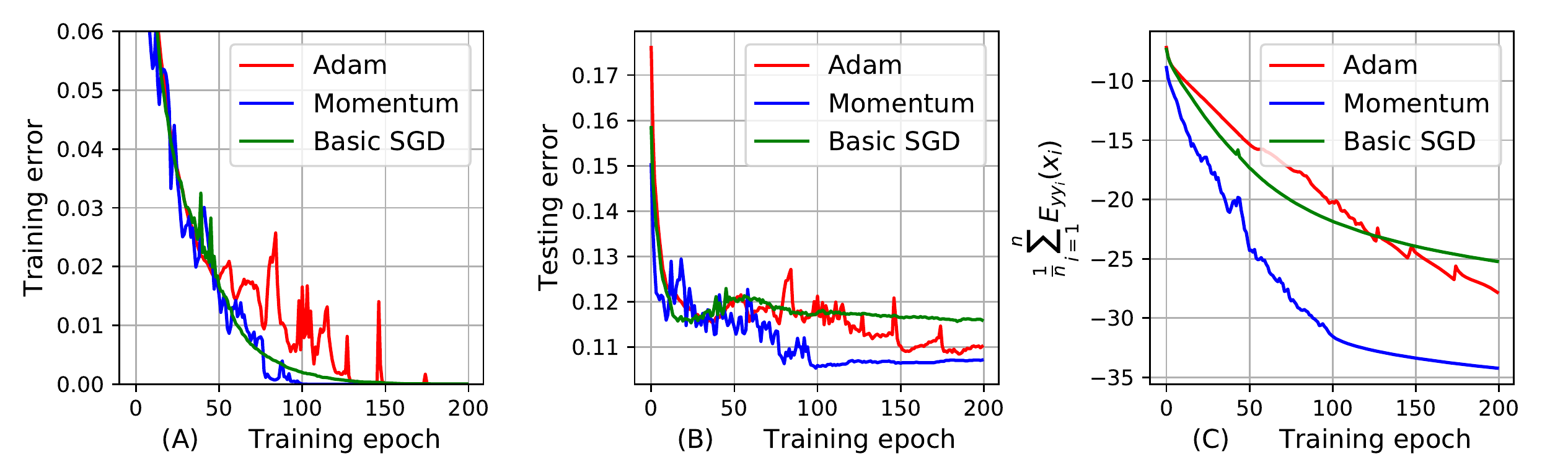}
\caption{
(A), (B), and (C) show the training error, the testing error, and the proposed PAC-Bayesian generalization bound over 200 training epochs, respectively. 
}
\label{Img_generalization_SGDs_A}
\end{figure}

We specify the $\text{MLP} = \{{x; f_1;f_2; f_Y}\}$ to classify the Fashion-MNIST dataset.
The MLP has 512 neurons in $f_1$ and 256 neurons in $f_2$.
All the activation functions are ReLU $\sigma(z):=\text{max}(0,z)$.
Since the dimension of a single Fashion-MNIST image is $28 \times 28$, $x$ has $M = 784$ input nodes.
Since the Fashion-MNIST dataset consists of $10$ different labels, ${f_Y}$ has $L = 10$ output nodes.
We still use three different training algorithms (the basic SGD, Momentum, and Adam) to train the MLP 200 epochs, and visualize the training error, the testing error, and the generalization bound in Figure \ref{Img_generalization_SGDs_A}.

We can observe that the bounds of the MLP with different training algorithms is consistent with the corresponding testing errors of the MLP.
In contrast to Adam achieving the lowest testing error for classifying the MNIST dataset (Figure \ref{Img_generalization_SGDs}), Momentum achieves the lowest testing error for classifying the Fashion-MNIST dataset. Figure \ref{Img_generalization_SGDs_A} shows the bound of the MLP with Momentum is also smaller than Adam and the basic SGD.

\pagebreak
\textbf{The bound can reflect the same sample size dependence as the testing error.}

\begin{figure}[htp]
\centering
\includegraphics[scale=0.45]{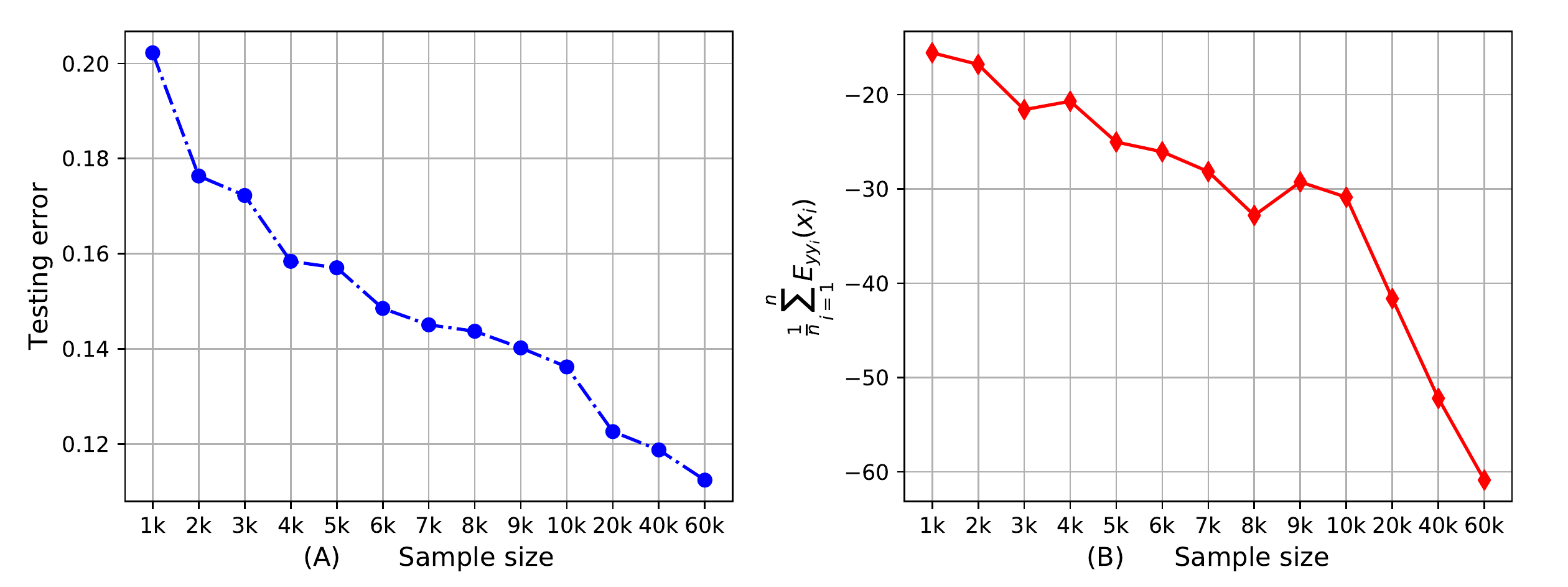}
\caption{
(A) and (B) show the testing error and the proposed PAC-Bayesian generalization bound given different training sample sizes, respectively. 
}
\label{Img_generalization_sample_size_A}
\end{figure}

Similar to the experiment in Section \ref{simulations}, we also create 13 different subsets of the Fashion-MNIST dataset with different number of training samples, and train the MLP2 designed in Section \ref{simulations} on the 13 subsets.
Figure \ref{Img_generalization_sample_size_A} shows the testing error and the bound with different sample sizes.
Similar to Figure \ref{Img_generalization_sample_size}, the bound shows a general decreasing trend as the training sample size increases, which is consistent with the variation of the testing error.

\textbf{The bounds can reflect the generalization performance of MLPs given random labels.}

\begin{figure}[htp]
\centering
\includegraphics[scale=0.45]{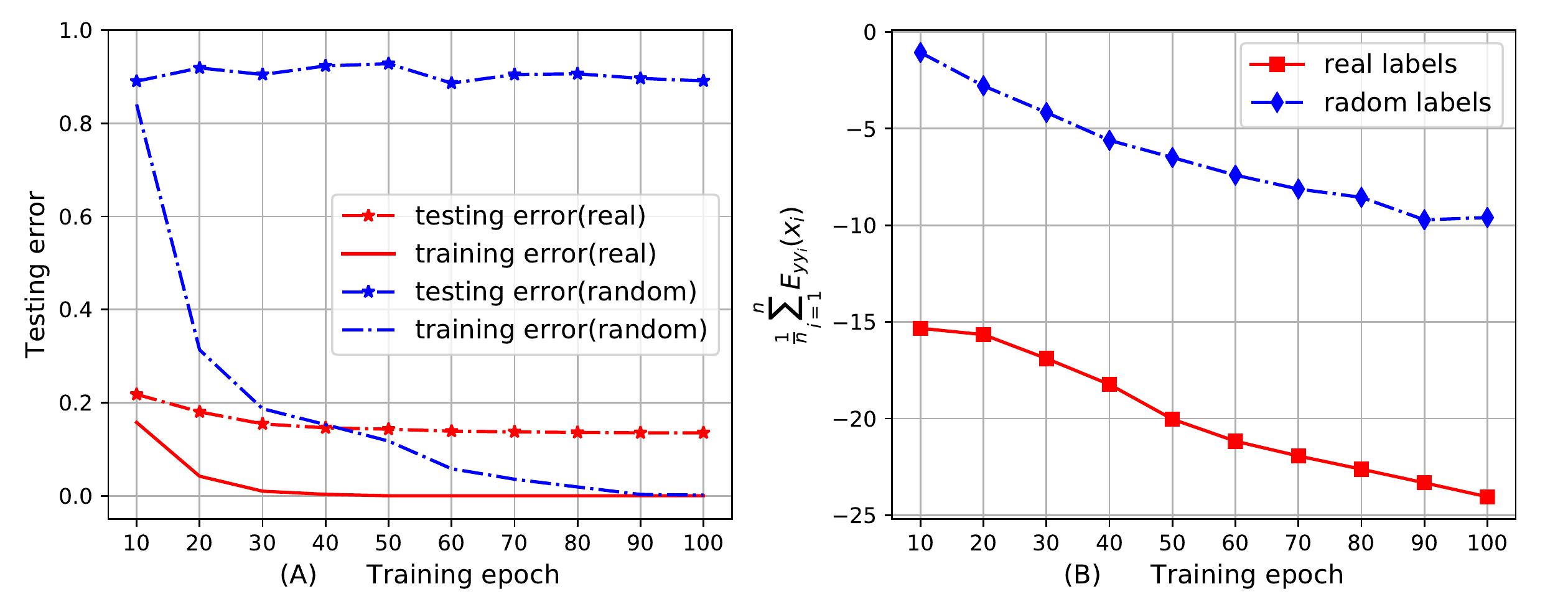}
\caption{
(A) and (B) show the testing error and the proposed PAC-Bayesian generalization bound given real labels and random labels, respectively. 
}
\label{Img_generalization_random_A}
\end{figure}

We train the MLP3 designed in Section \ref{simulations} 100 epochs to classify 10,000 samples of the Fashion-MNIST dataset with random labels, and visualize the training error, the testing error and the bound in Figure \ref{Img_generalization_random_A}.

We observe that the bound of MLP3 trained by random labels is higher than real labels, which is consistent with the testing error based on random labels is higher than real labels.

\end{document}